\documentclass{article}
\pdfoutput=1

\usepackage{microtype}
\usepackage{graphicx}
\usepackage{booktabs} 
\usepackage{tikz}
\usepackage{subcaption}
\usepackage{listings}
\usepackage{amsmath}
\usepackage{amsfonts}
\usepackage{mathtools}
\usepackage{todonotes}
\usepackage{my}
\usepackage{caption}
\usepackage{subcaption}
\usepackage{wrapfig}
\usepackage{qtree}
\usepackage{ulem}

\usepackage{tabularx}
\usepackage{courier}
\usepackage{xr-hyper}
\usepackage{syntax}
\usepackage[section]{placeins}
\usepackage[toc,page]{appendix}

\usepackage{hyperref}

\newcommand{\del}[1]{}
\newcommand{\new}[1]{#1}

\lstdefinestyle{base}{
  basicstyle=\fontsize{7}{10}\ttfamily,
  moredelim=**[is][\color{orange}]{@}{@},
  moredelim=**[is][\color{red}]{^}{^},
  moredelim=**[is][\color{violet}]{&}{&},
}

\pgfdeclareimage[height=0.6cm]{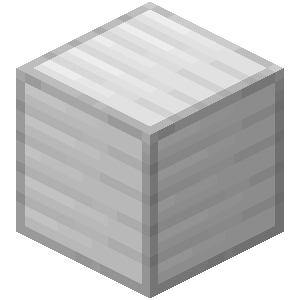}{figures/iron}
\pgfdeclareimage[height=0.6cm]{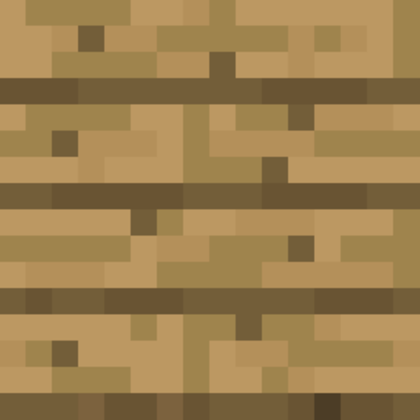}{figures/tree}
\pgfdeclareimage[height=0.6cm]{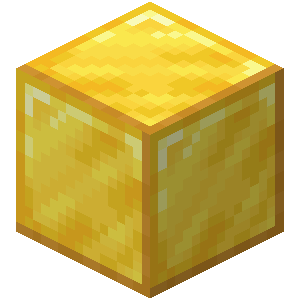}{figures/gold}
\pgfdeclareimage[height=0.6cm]{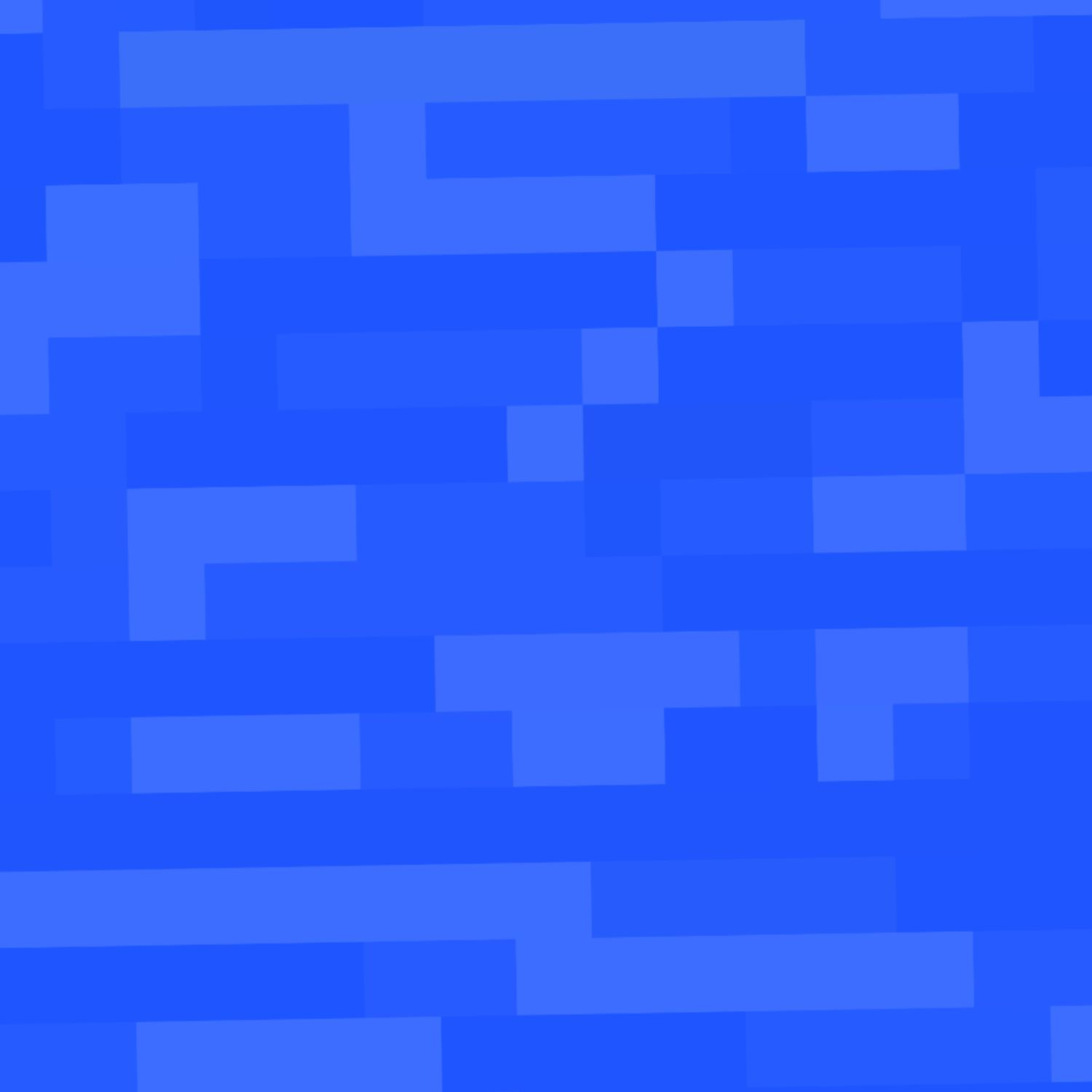}{figures/water}
\pgfdeclareimage[height=0.6cm]{bridge}{figures/wood}
\pgfdeclareimage[height=0.6cm]{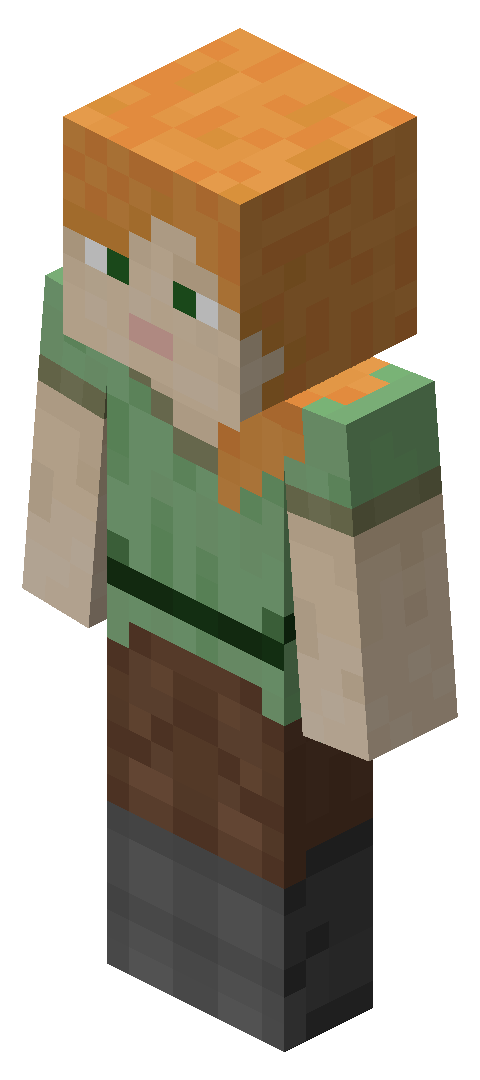}{figures/agent}
\pgfdeclareimage[height=0.6cm]{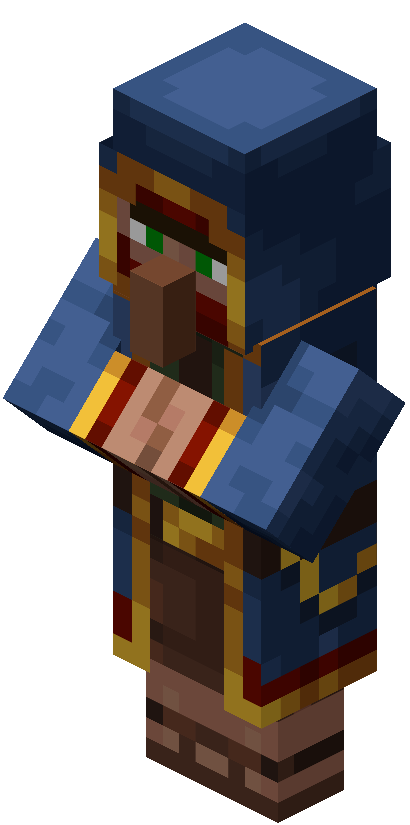}{figures/llama}
\pgfdeclareimage[height=0.6cm]{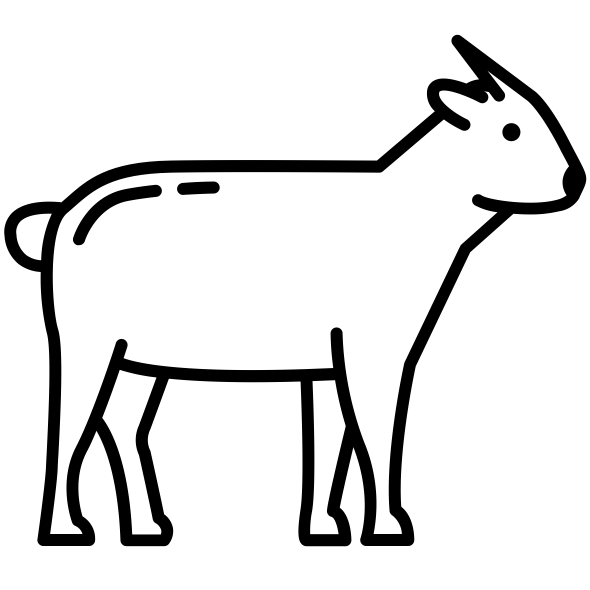}{figures/goat}
\pgfdeclareimage[height=0.6cm]{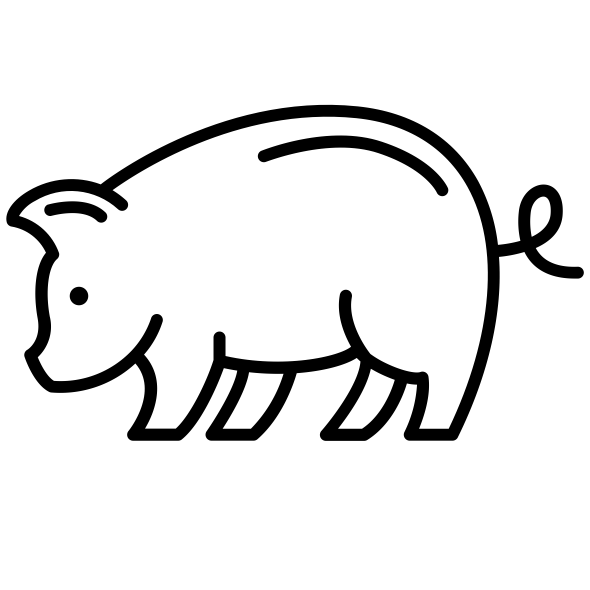}{figures/pig}
\pgfdeclareimage[height=0.6cm]{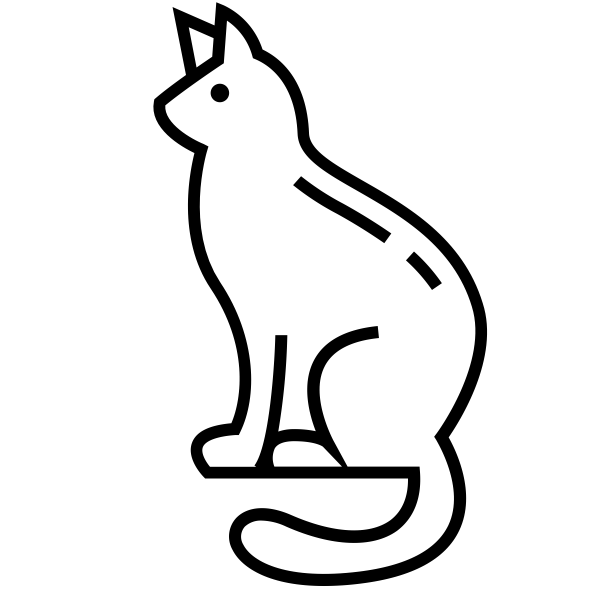}{figures/cat}
\pgfdeclareimage[height=0.6cm]{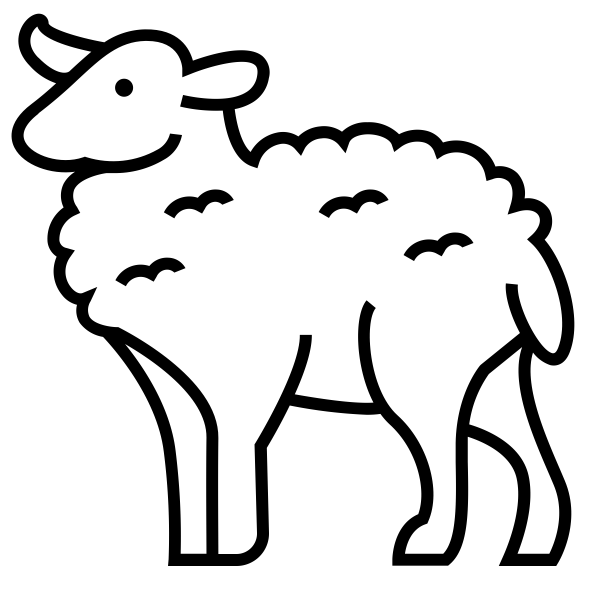}{figures/sheep}
\pgfdeclareimage[height=0.6cm]{agent}{figures/agent}
\pgfdeclareimage[height=0.6cm]{probe}{figures/probe}
\pgfdeclareimage[height=0.6cm]{minerals}{figures/minerals}
\pgfdeclareimage[height=0.6cm]{assimilator}{figures/assimilator}
\pgfdeclareimage[height=0.6cm]{nexus}{figures/nexus}
\pgfdeclareimage[height=0.6cm]{gateway}{figures/gateway}
\pgfdeclareimage[height=0.6cm]{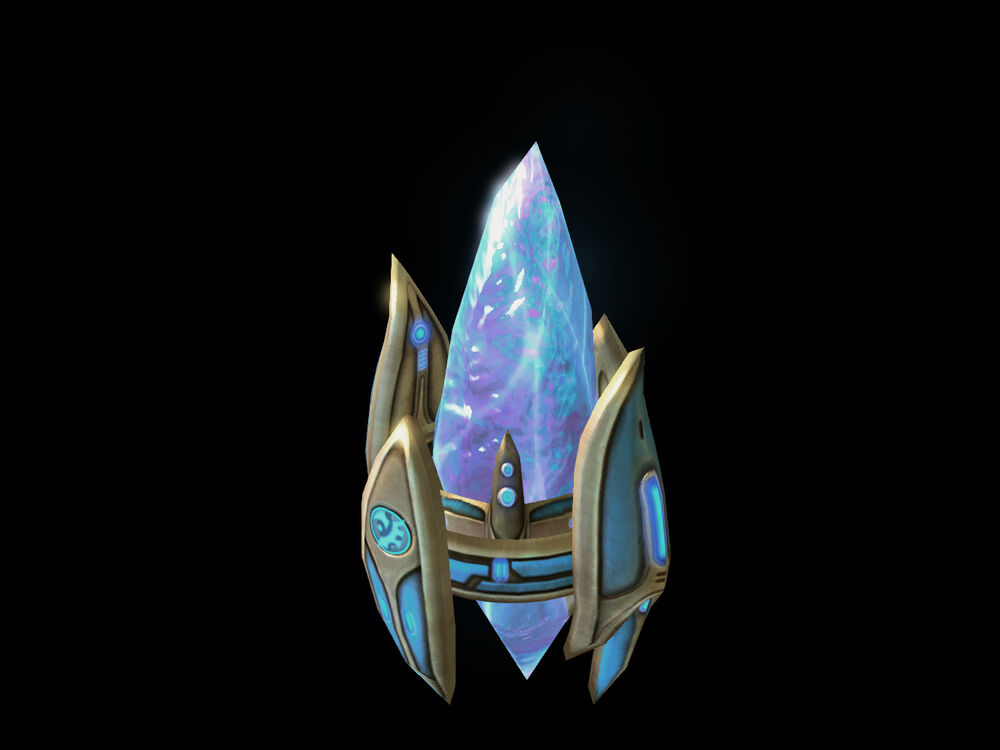}{figures/pylon}
\pgfdeclareimage[height=0.6cm]{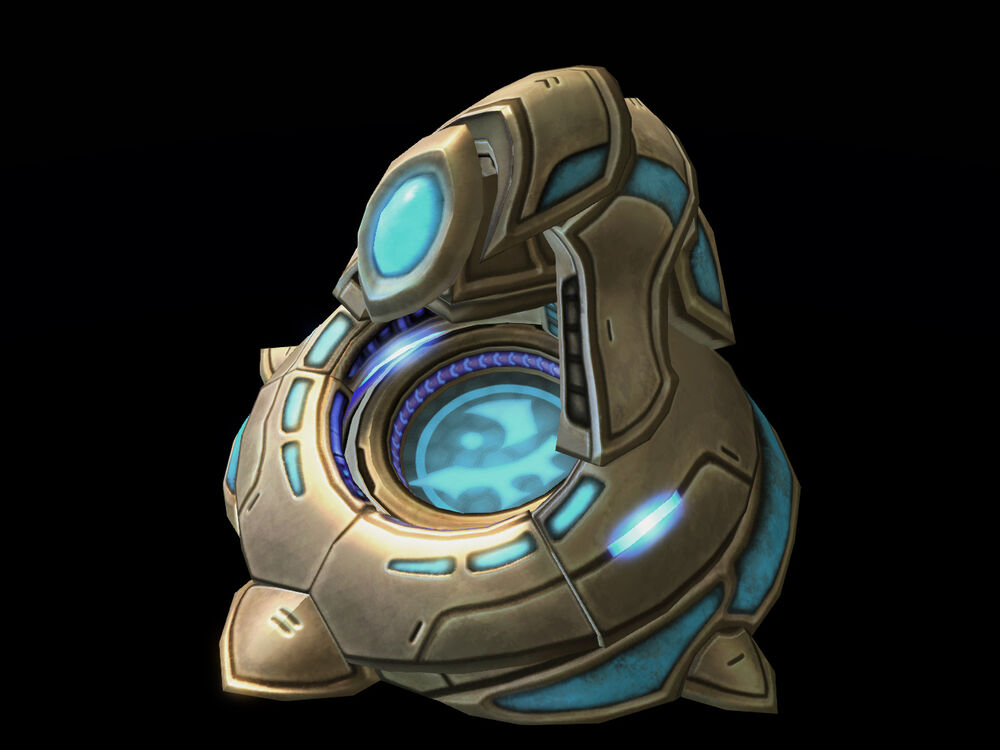}{figures/robotics-facility}

\usetikzlibrary{automata,positioning}
\usetikzlibrary{decorations.pathreplacing}
\usetikzlibrary{decorations.pathmorphing}
\usetikzlibrary{backgrounds}
\usetikzlibrary{arrows, chains, decorations.markings, shadows, shapes.arrows}
\usepackage[accepted]{icml2021}

\graphicspath{{figures/}} 

\makeatletter
\newcommand*{\addFileDependency}[1]{
  \typeout{(#1)}
  \@addtofilelist{#1}
  \IfFileExists{#1}{}{\typeout{No file #1.}}
}
\makeatother

\usetikzlibrary{decorations.pathreplacing}
\usetikzlibrary{decorations.pathmorphing}
\usetikzlibrary{snakes}
\newcommand{\highlight}[1]{\textcolor{red}{#1}}

\tikzset{snake it/.style={decorate, decoration=snake}}

\lstdefinestyle{base}{
  basicstyle=\fontsize{7}{10}\ttfamily,
  moredelim=**[is][\color{orange}]{@}{@},
  moredelim=**[is][\color{red}]{^}{^},
  moredelim=**[is][\color{violet}]{&}{&},
}

\icmltitlerunning{Reinforcement Learning of Implicit and Explicit Control Flow in Instructions}

\usepackage{subfiles} 

\begin{document}

\twocolumn[
    \icmltitle{Reinforcement Learning of Implicit and Explicit Control Flow in Instructions}



    \icmlsetsymbol{equal}{*}

    \begin{icmlauthorlist}
        \icmlauthor{Ethan A.~Brooks}{um}
        \icmlauthor{Janarthanan Rajendran}{um}
        \icmlauthor{Richard L.~Lewis}{wein}
        \icmlauthor{Satinder Singh}{um}
    \end{icmlauthorlist}
    \icmlaffiliation{um}{
        Department of Computer Science,
        University of Michigan}
    \icmlaffiliation{wein}{
        Weinberg Institute for Cognitive Science,
        Departments of Psychology and Linguistics,
        University of Michigan}

    \icmlcorrespondingauthor{Ethan Brooks}{ethanbro@umich.edu}

    \icmlkeywords{Machine Learning, ICML}

    \vskip 0.3in
]



\printAffiliationsAndNotice{}  

\begin{abstract}
    Learning to flexibly follow task instructions in dynamic environments poses interesting challenges for reinforcement learning agents. We focus here on the problem of learning control flow that deviates from a strict step-by-step execution of instructions—that is, control flow that may skip forward over parts of the instructions or return backward to previously completed or skipped steps. Demand for such flexible control arises in two fundamental ways: explicitly when control is specified in the instructions themselves (such as conditional branching and looping) and implicitly when stochastic environment dynamics require re-completion of instructions whose effects have been perturbed, or opportunistic skipping of instructions whose effects are already present. We formulate an attention-based architecture that meets these challenges by learning, from task reward only, to flexibly attend to and condition behavior on an internal encoding of the instructions. We test the architecture's ability to learn both explicit and implicit control in two illustrative domains---one inspired by Minecraft and the other by StarCraft---and show that the architecture exhibits zero-shot generalization to novel instructions of length greater than those in a training set, at a performance level unmatched by \del{two}\new{three} baseline recurrent architectures and one ablation architecture.
\end{abstract}
\normalem

\section{Introduction}
An important goal in artificial intelligence is developing flexible and
autonomous agents capable of accomplishing tasks that humans specify in
forms that are expressive to the agent and convenient for the human user. In this work
we focus on the reinforcement learning problem of following task instructions
that require the agent to learn control flow either because the instructions
themselves include explicit conditionals such as the if- and while- statements
familiar from programming languages, or because the need arises implicitly when
stochastic events in the environment require parts of the instructions to be
redone or allow them to be skipped. In a classical programming language
interpreter, the logic of control flow is fixed in advance and determined
completely by the program itself. In contrast, in our interactive \new{reinforcement-learning} (RL) setting,
while the reward is determined by the instructions, the control flow arises
dynamically from interactions between the instructions and the stochastic
dynamics of the environment.



Our primary contribution is a novel neural network architecture, the
Control Flow Comprehension Architecture - Scan (\name{}-S)\footnote{Source code may be accessed from \url{https://github.com/ethanabrooks/CoFCA-S}}, that successfully learns, from task
reward signals alone, how to follow instructions that require subtasks to be
performed in a nonlinear fashion. More specifically, we introduce two novel elements in our architecture that facilitate generalisation to novel and longer instructions during testing relative to during training: 1) an attentional mechanism that learns to maintain a pointer into the instructions and moves that pointer based on where the pointer should go next rather than how far the pointer should move, and 2) a mechanism that processes the instructions relative to the current pointer\del{ movement that captures}\new{, thereby capturing} a kind of coarse translation invariance present in our tasks.


We compare our method to \del{three}\new{four} baselines \new{and one ablation}, and demonstrate better performance on
two distinct domains, one designed to showcase explicit control-flow, and
the other to showcase implicit control-flow. We
argue that the improvements on these baselines are a result of our pointer
architecture, the freedom of pointer movement that it affords, and the
techniques that our architecture employs to transfer logic learned on shorter
instructions to instructions of much greater length.


\section{Related Work}

This section reviews prior work on neural architectures for \textit{instruction following}.
We also present a brief overview of work on \textit{program execution} because some of that work can handle explicit forms of control flow and because our learning architecture builds on architectures used in this line of work. However, our main \del{goal}\new{focus} in this paper is on methods that can learn\new{,} through trial-and-error interaction with an environment\new{,} the flow of control needed to obtain reward in the face of stochastic events in the environment both with and without explicit control flow in the instructions.


\paragraph{Neural architectures for instruction following.}
Our work is closely related to \textit{Zero-shot Task Generalization with Multi-Task Deep Reinforcement Learning} \cite{oh2017zero} (henceforth,
OLSK), which presents a novel architecture that accepts instructions in
the form of a sequence of symbols denoting actions and objects whose semantics
are learned with RL. The architecture maintains an attentional program pointer
in an approach similar to ours. An explicit analogy mechanism supports
generalizations to entirely novel action-object combinations, a kind of
generalization that is not the focus of our work. The instructions in OLSK
adhere to linear ordering\del{as described in the introduction}, \new{where tasks must be performed in exactly the same order that they occur in the instructions, }but the OLSK
architecture can be applied to our instructional task setting and we therefore
use it as an instructive baseline. The work of \citet{sun2020program}, whose
domain inspired the ``Minecraft'' domain in our work, considers task specifications
with rigid control-flow which their architecture navigates using a pre-specified
parser and program interpreter. They train a classifier to evaluate control-flow predicates using supervised learning---e.g., to determine whether an if-condition passes based on the number of some resource in the environment---
and an RL agent to interpret instructions based on observations of the environment and outputs of the classifier and parser.
The \textit{neural subtask graph solver} (NSGS) of \citet{sohn2018hierarchical}
accepts task specifications in the form of \textit{subtask graphs} encoding the
pre-condition dependencies among subtasks and rewards along the way; optimal
task execution requires finding graph traversals that maximize reward. The NSGS
is trained with RL and exhibits zero-shot generalization to novel graphs. The
graph specification is more expressive than our sequential instructions language
in that it makes explicit precondition dependencies, but it does not deal with the
non-linear control flow that is our concern here.

\new{
    Like our work, the preceding publications focus on instructions comprising sequences
    or collections of subtasks. A much broader literature has taken up the question of
    single-task instructions.
    \citet{yu2018guided} propose a novel deep architecture
    for combining the instruction sentence with an high-dimensional observation.
    \citet{yu2018interactive} extends this work to focus on generalization to instruction sentences
    comprising novel combinations of words (these describe new
    tasks, not new orderings of subtasks as in our case).
    \citet{bahdanau2018learning} use an adversarial framework to generate reward functions from instructions,
    thereby reducing the need for hand-engineering.
}

\new{
    Another important body of literature considers human-generated natural language instructions
    (as opposed
    to the programmatically generated instructions in our domains). In general, these approaches use supervised learning or consider instructions that induce simpler policies than those required by our domain.
    \citet{chaplot2018gated} propose a
    multiplication-based mechanism for combing instructions with observations,
    which facilitates following of natural-language instructions in 3d domains.
    \citet{misra2018mapping} develop
    a supervised-learning approach for mapping natural language statements to sequences of actions
    by marginalizing across possible goal locations implied by the
    instruction. \citet{fried2018speaker} consider a similar problem, but introduce a ``speaker''
    network that infers the probability of instructions given action trajectories. They use the ``speaker''
    to augment the dataset with synthetic instructions and to rank candidate action-sequences
    produced by a ``follower'' network.
    \citet{hill2020human} use embeddings from a pre-trained language model to
    encode a single-sentence instruction. The result is fed into an RL agent for
    execution in a simulated 3d-domain.
}

\paragraph{Neural architectures for program execution.}
\citet{graves2014neural} was an important early contribution to the problem of program
execution using end-to-end differentiable architectures.
Our work builds on the seminal attention-based
architecture from this paper but as we detail below we needed a more sophisticated attention mechanism to allow for learning of the kinds of nonlinear control flow required in our reinforcement learning tasks.
\citet{reed2015neural}
develops an architecture that learns to execute complex programs via a
supervised training regime that paired programs and execution traces.
This work
does not consider the problem of following  instructions with many
parts whose ordering must be learned from trial-and-error experience.
\citet{bovsnjak2017programming} extend this work with supervised learning of tasks specified using the Forth programming language.
\citet{bieber2020learning} develop an architecture that
predicts the output of a program.


Our work thus exists at the intersection of work that learns to follow
instructions comprising multiple subtasks---\cite{oh2017zero, sun2020program, sohn2018hierarchical}---and work that focuses on
the problem of non-linear task specifications---\cite{bovsnjak2017programming, bieber2020learning}. Our work departs from prior work in several ways: 1) unlike the work on program execution/interpretation, in our setting it is not possible to determine the sequence of subtasks to be performed solely by looking at the instructions, rather it has to be learned through trial and error interaction with a stochastic environment, and 2) we are in the RL setting in which the only feedback to the agent is in the form of a terminal (and therefore delayed) reward as opposed to much of the prior work that has focused on variations within the supervised learning setting.


\section{
  \name{}-S: The Control Flow Comprehension Architecture - Scan} 

\label{approach}

In our setting, it is not possible to map the instructions and current
observation to the desired action because the current observation may not record
all the subtasks that have already been performed and those that have been
undone by stochastic events. One way to deal with this would be to learn a
mapping from the instruction and the history of interactions to the action.
However this can be challenging in complex settings with long instructions where
each line requires the agent to take several actions in the environment. The
underlying assumption in our architecture is that much of this challenge can be
handled by maintaining a pointer to a line/subtask in the instructions, and one
can learn a mapping from the line/subtask pointed to by the pointer and the
current observation to a desired action. Of course, this requires a procedure
for updating the pointer after taking the action and receiving an observation.

One way to handle pointer movement would be to learn a distribution over
backward and forward pointer movements during training. The challenge is that
one can only learn about movement distances encountered during training. \del{Our}\new{One of our}
main contribution\new{s} is to learn \textit{where} we want to move the pointer instead
of \textit{how far} we want to move the pointer. The \new{``Scan''} mechanism we introduce to do
this \del{is to}compute\new{s} the probability of a pointer movement as a function of the
features of the lines around the line to which we are considering a move. This allows
pointer movement to be free of distances and allows generalization to longer
instructions requiring longer distance pointer movements during testing than
seen during training.

Furthermore we introduce a mechanism in our architecture to exploit the following invariance property present in our tasks: chunks of the instructions, e.g., a while loop, require the agent to take the same actions no matter where that chunk is in the instructions. We accomplish this by always processing the embeddings of the instructions from the pointer's position in our architecture. This simple mechanism captures the chunk-translation-invariance property and facilitates transfer of learned behavior from one position in the instructions to other positions in the instructions. This contributes to our results on generalising to longer instructions during testing.

\begin{figure}[t]
    \centering
    \begin{subfigure}[t]{.44\columnwidth}
        \resizebox*{\textwidth}{!}{
            \begin{tikzpicture}[x=2mm,y=2mm,
                    bin/.style={
                            rectangle,
                            minimum height=8mm,
                            thin, draw,
                        },
                    weight/.style={
                            rectangle,
                            minimum height=1mm,
                            thin, draw,
                        },
                    weights/.style={
                            rectangle,
                            minimum height=1mm,
                            minimum width=24mm,
                            thin, draw,
                        },
                    memory/.style={
                            rectangle,
                            minimum height=8mm,
                            minimum width=24mm,
                            thin, draw,
                        },
                ]
                \node[memory,label={[xshift=-19.0mm]right:Memory $\mem$}]
                (M) at (-3, 2) {};
                \node[memory]                         (P) at (5, 22) {$\edges$};
                \node at (-7, -1) {$\ptr{t}$};
                \node[bin]                            (Mp) at (-7, 2) {};
                \node[minimum size=1cm, draw,circle]  (x) at (6, 2) {$\obs{t}$};
                \node[draw,circle,minimum size=1cm]   (edgeNet) at (5, 10) {$\edgeNet$};
                \node[draw,circle,minimum size=1cm]   (upperPolicy) at (-5, 10) {$\policy$};
                \fill [black!5]                      (-1,16) rectangle (0.5,18);
                \fill [black!10]                      (0.5,16) rectangle (2,18);
                \fill [black!20]                      (2,16) rectangle (3.5,18);
                \fill [black!10]                      (3.5,16) rectangle (5,18);
                \fill [black!5]                      (5,16) rectangle (6.5,18);
                \draw [decorate,decoration={brace}]
                (-1,18.5) -- (11,18.5) node[midway,yshift=2mm]{\tiny softmax};
                \node[weights]                        (u) at (5, 17) {$\edgeChoice{t}$};
                \node                                 (p1) at (5, 30) {$\ptrDelta{t}$};
                \node                                 (g) at (-5, 17) {$\action{t}$};

                \draw [decorate,decoration={brace}]
                (-1,24.5) -- (11,24.5) node[midway,yshift=2mm]{};
                \node   (weighted-sum) at (5, 25.5) {\tiny weighted-sum};

                \draw [->] (Mp)          to [out=90,in=225] (edgeNet);
                \draw [->] (Mp)          to [out=90,in=250] (upperPolicy);
                \draw [->] (x)          to  [out=90, in=285] (edgeNet);
                \draw [->] (x)          to  [out=90,in=305] (upperPolicy);
                \draw [->] (upperPolicy) to  node[left] {\tiny sample} (g);
                \draw [->] (edgeNet) to (u);
                \draw[->] (weighted-sum) -> node[right] {\tiny sample} (p1);

            \end{tikzpicture}
        }
    \end{subfigure}
    \hfill
    \begin{subfigure}[b]{.5\columnwidth}
        \begin{algorithm}[H]
            \footnotesize
            \caption*{\name{}-S Algorithm}
            \begin{algorithmic}[1]
                \REQUIRE pointer $\ptr{t}$
                \REQUIRE observation $\obs{t}$
                \REQUIRE instruction $\task$
                \STATE $\mem \gets \bagOfWords{\task}$
                \STATE $\bigruOutputi{t} \gets \bigruOf{[\mem, \obs{t}]}$
                \COMMENT{BI-GRU starts at line $\ptr{t}$}
                \STATE Generate $\edges$ (equation $\ref{eq:P})$
                \STATE $\action{t} \sim \policyOf{\obs{t}, \memi{\ptr{t}}}$
                \STATE $\edgeChoice{t}  \gets \edgeNetOf{\obs{t},
                        \memi{\ptr{t}}, \action{t}}$
                \STATE $\edgeChoiceSoftmax{t} \gets \softmax{\edgeChoice{t}}$
                \STATE $\ptrDelta{t} \sim \cat{\edges\edgeChoiceSoftmax{t}}$
                \STATE $\gate{t} \sim \gateNetOf{\obs{t}, \memi{\ptr{t}}, \action{t}}$
                \STATE $\ptr{t + 1} \gets \ptr{t} + \gate{t}\ptrDelta{t}$
            \end{algorithmic}
        \end{algorithm}
    \end{subfigure}
    \caption{(Left) Depicts the flow of
        information every time step from memory $\mem$, pointer $\ptr{t}$, and
        observation $\obs{t}$ to actions $\action{t}$ and pointer movements
        $\ptrDelta{t}$. (Right) Pointer update pseudocode. }
    \label{schematic}
\end{figure}

Our architecture comprises the following primary components. $\mem$ is an encoding of the instructions.
$\ptr{t}$ is the integer pointer into $\mem$. $\policy$, implemented as a neural network, combines information from $\mem$ and the observation, $\obs{t}$ to produce actions. $\edges$ is a collection of possible pointer movement distributions. $\edgeNet{}$, implemented as a neural network, combines information from $\mem$ and $\obs{t}$ to choose among these distributions.  Fig.~\ref{schematic} identifies these components and their relationships. We now describe them in detail.

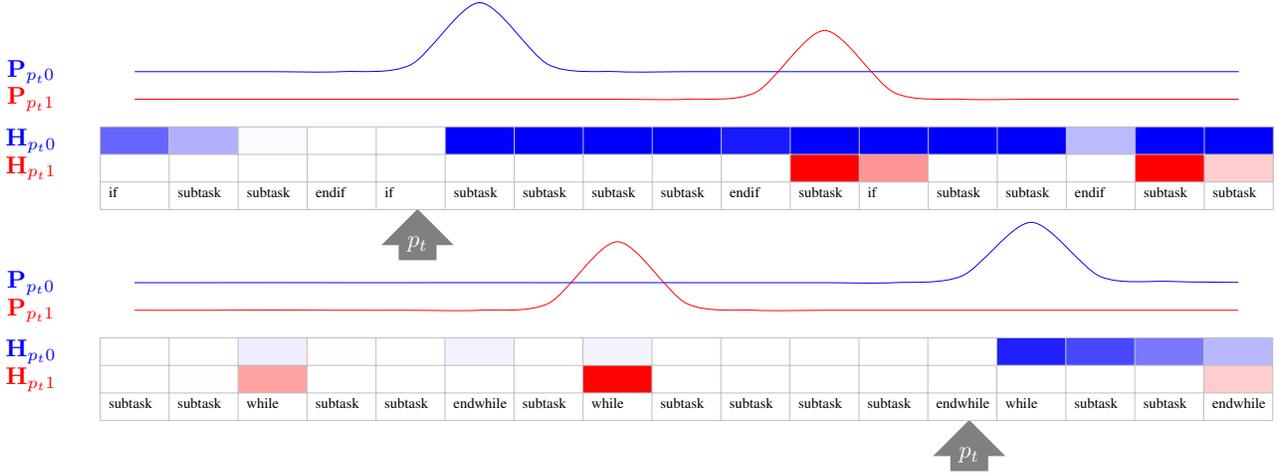
\begin{figure*}[ht]
    \begin{subfigure}[t]{\textwidth}
        \centering
        \resizebox*{\textwidth}{!}{
            \begin{tikzpicture}[
                    line/.style={font=\tiny,anchor=west, yshift=0.25cm},
                    H1/.style={xshift=0.5cm,yshift=1.00cm,minimum width=1cm,minimum height=0.4cm},
                    H2/.style={xshift=0.5cm,yshift=0.60cm,minimum width=1cm,minimum height=0.4cm}
                ]
                \newcounter{x}
                \setcounter{x}{0}
                \draw (\value{x},0) node[H1,fill=blue!60.0766168] {};
                \draw (\value{x},0) node[H2,fill=red!0.08448899] {};
                \draw (\value{x},0) node[line] {if};
                \stepcounter{x}
                \draw (\value{x},0) node[H1,fill=blue!30.430737] {};
                \draw (\value{x},0) node[H2,fill=red!0] {};
                \draw (\value{x},0) node[line] {subtask};
                \stepcounter{x}
                \draw (\value{x},0) node[H1,fill=blue!2.0080663] {};
                \draw (\value{x},0) node[H2,fill=red!0] {};
                \draw (\value{x},0) node[line] {subtask};
                \stepcounter{x}
                \draw (\value{x},0) node[H1,fill=blue!0.027454994] {};
                \draw (\value{x},0) node[H2,fill=red!0] {};
                \draw (\value{x},0) node[line] {endif};
                \stepcounter{x}
                \draw (\value{x},0) node[H1,fill=blue!0] {};
                \draw (\value{x},0) node[H2,fill=red!0] {};
                \draw (\value{x},0) node[line] {if};
                \stepcounter{x}
                \draw (\value{x},0) node[H1,fill=blue!99.9826] {};
                \draw (\value{x},0) node[H2,fill=red!0] {};
                \draw (\value{x},0) node[line] {subtask};
                \stepcounter{x}
                \draw (\value{x},0) node[H1,fill=blue!99.996436] {};
                \draw (\value{x},0) node[H2,fill=red!0] {};
                \draw (\value{x},0) node[line] {subtask};
                \stepcounter{x}
                \draw (\value{x},0) node[H1,fill=blue!99.99975] {};
                \draw (\value{x},0) node[H2,fill=red!0] {};
                \draw (\value{x},0) node[line] {subtask};
                \stepcounter{x}
                \draw (\value{x},0) node[H1,fill=blue!99.99982] {};
                \draw (\value{x},0) node[H2,fill=red!0] {};
                \draw (\value{x},0) node[line] {subtask};
                \stepcounter{x}
                \draw (\value{x},0) node[H1,fill=blue!89.711285] {};
                \draw (\value{x},0) node[H2,fill=red!0] {};
                \draw (\value{x},0) node[line] {endif};
                \stepcounter{x}
                \draw (\value{x},0) node[H1,fill=blue!100] {};
                \draw (\value{x},0) node[H2,fill=red!99.99212] {};
                \draw (\value{x},0) node[line] {subtask};
                \stepcounter{x}
                \draw (\value{x},0) node[H1,fill=blue!99.998677] {};
                \draw (\value{x},0) node[H2,fill=red!41.971487] {};
                \draw (\value{x},0) node[line] {if};
                \stepcounter{x}
                \draw (\value{x},0) node[H1,fill=blue!99.9997] {};
                \draw (\value{x},0) node[H2,fill=red!0.4874527] {};
                \draw (\value{x},0) node[line] {subtask};
                \stepcounter{x}
                \draw (\value{x},0) node[H1,fill=blue!99.99659] {};
                \draw (\value{x},0) node[H2,fill=red!0.7373584] {};
                \draw (\value{x},0) node[line] {subtask};
                \stepcounter{x}
                \draw (\value{x},0) node[H1,fill=blue!27.167922] {};
                \draw (\value{x},0) node[H2,fill=red!0] {};
                \draw (\value{x},0) node[line] {endif};
                \stepcounter{x}
                \draw (\value{x},0) node[H1,fill=blue!100] {};
                \draw (\value{x},0) node[H2,fill=red!99.999213] {};
                \draw (\value{x},0) node[line] {subtask};
                \stepcounter{x}
                \draw (\value{x},0) node[H1,fill=blue!99.99976] {};
                \draw (\value{x},0) node[H2,fill=red!19.52194] {};
                \draw (\value{x},0) node[line] {subtask};
                \stepcounter{x}

                \draw[blue,transform canvas={yshift=2.0cm, xshift=.5cm}]  plot [smooth] coordinates {
                        (0,0)
                        (1,0)
                        (2,0)
                        (3,0.00027454994)
                        (4,0.1)
                        (5,0.99955153)
                        (6,0.10017043932)
                        (7,0)
                        (8,0)
                        (9,0)
                        (10,0)
                        (11,0)
                        (12,0)
                        (13,0)
                        (14,0)
                        (15,0)
                        (16,0)
                    };

                \draw[red,transform canvas={yshift=1.6cm, xshift=.5cm}]  plot [smooth] coordinates {
                        (0,0.0008448045)
                        (1,0)
                        (2,0)
                        (3,0)
                        (4,0)
                        (5,0)
                        (6,0)
                        (7,0)
                        (8,0)
                        (9,0.1)
                        (10,0.99684405)
                        (11,0.1)
                        (12,0)
                        (13,0)
                        (14,0)
                        (15,0)
                        (16,0)
                    };
                \draw (-1.5,0) node[H1, blue] {$\bigruOutputi{\ptr{t}0}$};
                \draw (-1.5,0) node[H2, red] {$\bigruOutputi{\ptr{t}1}$};
                \draw (-1.5,1) node[H1, blue] {$\edgesi{\ptr{t}0}$};
                \draw (-1.5,1) node[H2, red] {$\edgesi{\ptr{t}1}$};
                \node [single arrow, fill=gray, minimum height=.05cm,shape border rotate=90, text=white]            at (4.6, -0.5) {$\ptr{t}$};
                \draw[lightgray] (0,0) grid[xstep=1,ystep=.4] (17,1.2);
            \end{tikzpicture}
        }
    \end{subfigure}%
    \\
    \begin{subfigure}[b]{\textwidth}
        \centering
        \resizebox*{\textwidth}{!}{
            \begin{tikzpicture}[
                    line/.style={font=\tiny,anchor=west, yshift=0.25cm},
                    H1/.style={xshift=0.5cm,yshift=1.00cm,minimum width=1cm,minimum height=0.4cm},
                    H2/.style={xshift=0.5cm,yshift=0.60cm,minimum width=1cm,minimum height=0.4cm}
                ]
                \setcounter{x}{0}
                \draw (\value{x},0) node[H1,fill=blue!0.05921151] {};
                \draw (\value{x},0) node[H2,fill=red!0.014426647] {};
                \draw (\value{x},0) node[line] {subtask};
                \stepcounter{x}
                \draw (\value{x},0) node[H1,fill=blue!0.054720015] {};
                \draw (\value{x},0) node[H2,fill=red!0.8423797] {};
                \draw (\value{x},0) node[line] {subtask};
                \stepcounter{x}
                \draw (\value{x},0) node[H1,fill=blue!06.7415364] {};
                \draw (\value{x},0) node[H2,fill=red!036.613148] {};
                \draw (\value{x},0) node[line] {while};
                \stepcounter{x}
                \draw (\value{x},0) node[H1,fill=blue!0.2003259] {};
                \draw (\value{x},0) node[H2,fill=red!0] {};
                \draw (\value{x},0) node[line] {subtask};
                \stepcounter{x}
                \draw (\value{x},0) node[H1,fill=blue!0] {};
                \draw (\value{x},0) node[H2,fill=red!0] {};
                \draw (\value{x},0) node[line] {subtask};
                \stepcounter{x}
                \draw (\value{x},0) node[H1,fill=blue!004.7023594] {};
                \draw (\value{x},0) node[H2,fill=red!0] {};
                \draw (\value{x},0) node[line] {endwhile};
                \stepcounter{x}
                \draw (\value{x},0) node[H1,fill=blue!0] {};
                \draw (\value{x},0) node[H2,fill=red!000.1741815] {};
                \draw (\value{x},0) node[line] {subtask};
                \stepcounter{x}
                \draw (\value{x},0) node[H1,fill=blue!004.15618] {};
                \draw (\value{x},0) node[H2,fill=red!099.33696] {};
                \draw (\value{x},0) node[line] {while};
                \stepcounter{x}
                \draw (\value{x},0) node[H1,fill=blue!0.085231365] {};
                \draw (\value{x},0) node[H2,fill=red!0] {};
                \draw (\value{x},0) node[line] {subtask};
                \stepcounter{x}
                \draw (\value{x},0) node[H1,fill=blue!0.030979072] {};
                \draw (\value{x},0) node[H2,fill=red!0] {};
                \draw (\value{x},0) node[line] {subtask};
                \stepcounter{x}
                \draw (\value{x},0) node[H1,fill=blue!0] {};
                \draw (\value{x},0) node[H2,fill=red!0] {};
                \draw (\value{x},0) node[line] {subtask};
                \stepcounter{x}
                \draw (\value{x},0) node[H1,fill=blue!0] {};
                \draw (\value{x},0) node[H2,fill=red!0] {};
                \draw (\value{x},0) node[line] {subtask};
                \stepcounter{x}
                \draw (\value{x},0) node[H1,fill=blue!0] {};
                \draw (\value{x},0) node[H2,fill=red!0] {};
                \draw (\value{x},0) node[line] {endwhile};
                \stepcounter{x}
                \draw (\value{x},0) node[H1,fill=blue!87.5448] {};
                \draw (\value{x},0) node[H2,fill=red!0] {};
                \draw (\value{x},0) node[line] {while};
                \stepcounter{x}
                \draw (\value{x},0) node[H1,fill=blue!71.713614] {};
                \draw (\value{x},0) node[H2,fill=red!0] {};
                \draw (\value{x},0) node[line] {subtask};
                \stepcounter{x}
                \draw (\value{x},0) node[H1,fill=blue!52.647656] {};
                \draw (\value{x},0) node[H2,fill=red!0] {};
                \draw (\value{x},0) node[line] {subtask};
                \stepcounter{x}
                \draw (\value{x},0) node[H1,fill=blue!27.853474] {};
                \draw (\value{x},0) node[H2,fill=red!19.52194] {};
                \draw (\value{x},0) node[line] {endwhile};
                \stepcounter{x}

                \draw[blue,transform canvas={yshift=2.0cm, xshift=.5cm}]  plot [smooth] coordinates {
                        (0,0)
                        (1,0)
                        (2,0.00073248026)
                        (3,0)
                        (4,0)
                        (5,0.00054034067)
                        (6,0)
                        (7,0.0004996522)
                        (8,0)
                        (9,0)
                        (10,0)
                        (11,0)
                        (12,0.1)
                        (13,0.87544525)
                        (14,0.08931935)
                        (15,0.018542398)
                        (16,0.004641276)
                    };

                \draw[red,transform canvas={yshift=1.6cm, xshift=.5cm}]  plot [smooth] coordinates {
                        (0,0)
                        (1,0)
                        (2,0.0024199553)
                        (3,0)
                        (4,0)
                        (5,0)
                        (6,0.1)
                        (7,0.9932129)
                        (8,0.1)
                        (9,0)
                        (10,0)
                        (11,0)
                        (12,0)
                        (13,0)
                        (14,0)
                        (15,0)
                        (16,0)
                    };
                \draw (-1.5,0) node[H1, blue] {$\bigruOutputi{\ptr{t}0}$};
                \draw (-1.5,0) node[H2, red] {$\bigruOutputi{\ptr{t}1}$};
                \draw (-1.5,1) node[H1, blue] {$\edgesi{\ptr{t}0}$};
                \draw (-1.5,1) node[H2, red] {$\edgesi{\ptr{t}1}$};
                \node [single arrow, fill=gray, minimum height=.05cm,shape border rotate=90, text=white]            at (12.6, -0.5) {$\ptr{t}$};
                \draw[lightgray] (0,0) grid[xstep=1,ystep=.4] (17,1.2);
            \end{tikzpicture}
        }
    \end{subfigure}%
    \caption{
        (Upper) Here, $\ptr{t}=4$. The agent will choose the distribution
        $\edgesi{\ptr{t}0}$ if the condition succeeds and
        $\edgesi{\ptr{t}1}$  if it fails. To
        generate these distributions $\bigruOutputi{\ptr{t}0}$ has
        ``flagged''
        all lines except those immediately preceding $\ptr{t}$.
        $\bigruOutputi{\ptr{t}1}$ has flagged lines following \emph{endif}.     (Lower) Here, $\ptr{t}=12$. The agent will use the
        $\edgesi{\ptr{t}1}$ distribution to return the pointer to the start of
        the while loop in order to inspect the \emph{while} condition, whose evaluation
        will determine the next distribution that the agent chooses (as in the upper figure).
    }
    \label{fig:pointer-movement}
\end{figure*}

\subsection{Instruction preprocessing.}
\label{instruction-preprocessing}
First, our architecture uses a lookup table to embed each line of the instruction (line 1 of the pseudocode in Fig.~\ref{schematic}). If an instruction line comprises several symbols, we embed each symbol separately and sum them. The result is $\mem \in \realspace{\instructionLength\times \embeddingDim}$, where
$\instructionLength$ is the \del{length of}\new{number of lines in} the instructions and $\embeddingDim$ the
size of the embedding.
Next, we \del{pass $\mem$}\new{concatenate $\obs{t}$ to each line of $\mem$} and pass the result through a bidirectional \new{Gated Recurrent Unit} (GRU) \cite{chung2014empirical}\new{, a variant of the bidirectional Recurrent Neural Network \cite{graves2013hybrid}.}\del{, with the forward pass starting} \new{Importantly, we start the forward pass} at line $\ptr{t}$ (recall that $\ptr{t}$ is the pointer into memory)\new{, instead of the first line of the instruction. This simple change enables the architecture to exploit the chunk-translation-invariance property, by allowing learning from earlier lines of the instruction to be reused on later lines.} \del{The GRU's embedding assigns $\numEdges$ weights to }\new{The GRU outputs an $\numEdges$ dimensional vector for} each line, \del{where}\new{with} higher weights \del{increase}\new{increasing} the probability of moving forward to that line (more explanation in the next paragraph). We do the same with a backward GRU starting at line $\ptr{t}-1$, this time for backward movement. Combining the two we get a weight matrix $\bigruOutputi{t} \in \realspace{2\instructionLength \times \numEdges}$.
$\numEdges$  encodes the number of possible pointer distributions which we subsequently choose among using information derived from the observation. We pass $\bigruOutputi{t}$ through a sigmoid function squashing it between 0 and 1. \new{These steps correspond to line 2 of the pseudocode.}

\subsection{The Scan Mechanism.}
\label{scan}
\new{This mechanism transforms  $\bigruOutputi{t}$ into a collection of distributions over pointer movements.} To simplify our explanation of \del{the eponymous ``scanning''}\new{this} mechanism\del{, which produces the pointer movement distribution}, we first assume that $\numEdges = 1$. In this case,
the distribution we use to generate pointer movements is equivalent to a geometric distribution generated by the following process. We scan through each line in the order $\ptr{t} + 1, \ptr{t} - 1, \ptr{t} +2, \ptr{t} -2, \dots$.  At each line we flip a coin, with heads-probability determined by the sigmoid output described in the preceeding paragraph. We stop at the first line where the coin comes up heads. When $\numEdges > 1$, we repeat this process $\numEdges$ times. The result is the matrix

\begin{align}
    \label{eq:P}
    \edgesi{ij}    & = \sigmoidOf{\bigruOutputi{ij}}\;\;\underset{\smash{\mathclap{k \in \set{1, -1, 2, -2,
                    \dots, \instructionLength, -\instructionLength},k\ne i }}}{\prod}\;\;\left(1 -
    \sigmoidOf{\bigruOutputi{(\instructionLength + k) j}}\right)
    \\
    \intertext{\new{Like $\bigruOutputi{t}$, $\edges$ is in $\realspace{2\instructionLength \times \numEdges}$.} We convert $\edges$ into a single distribution using $\edgeChoice{t}$, weighting over the different distributions in $\edges$, as follows:}
    \edgeChoice{t} & = \edgeNetOf{\obs{t}, \memi{\ptr{t}}, \embedding{\action{t}}}
    \\
    \ptrDelta{t}   & \sim \sum_{i=1}^\numEdges \softmax{\edgeChoice{t}}_i \edges_{(\cdot) i}
\end{align}

where $\edgeNet$ is a neural network (details in \S\ref{network-architectures})
and $\ptrDelta{t}$ is a pointer movement in the form of a delta to add to
$\ptr{t}$.
These steps correspond to lines 5 through 7 of the pseudocode.

The motivation for the distribution expressed in equation \ref{eq:P} is that it allows the size of pointer
movements to depend on features at the destination line, not on the size of the
jump. This is critical to enable the agent to perform pointer movements larger than those performed during training. E.g., if $\bigruOutput_{0}$ corresponds to an \emph{if} line, the
GRU might learn to flag subsequent \emph{endif} lines by assigning large
values to $\bigruOutputi{ij}$ for all indices $i$ corresponding to subsequent
\emph{endif} lines.
As long as all values of $\bigruOutput_{kj}$ are nearly 0 for $\abs{k} < \abs{i}$,
the pointer will be able to move to line $i$, even if $i$ is much larger than any jump
that the agent has performed during training.

\subsection{Gating of pointer movement.}
\label{gating}
While performing an individual subtask, the agent should not move the instruction
memory pointer---it should learn to wait for the subtask to be completed before
advancing. The agent learns to \emph{gate}  changes to the memory pointer
to accomplish this waiting.
The gate is a binary value $\gate{t}$ sampled from a learned distribution
$\gateNetOf{\obs{t}, \memi{\ptr{t}}, \embedding{\action{t}}}$.
$\gateNet$ is a feed-forward neural network as detailed in \S \ref{network-architectures}
.
$\ptr{t}$ only changes position when the gate's value is 1:
$ \ptr{t+1} = \ptr{t} + \gate{t}\ptrDelta{t}$. This corresponds to lines 8 and 9 of the pseudocode.

\subsection{Action sampling mechanism.}
\label{action-sampling}
Actions depend on information from the instruction and from the environment. In
one of our domains, actions also depend on the history of previous actions. We
derive information from the instruction by using the pointer to index into the
encoded representation of the instruction, $\mem_{\ptr{t}}$. We derive
information about the environment from an encoding of the current observation
$\obs{t}$ (see \S\ref{network-architectures}). Where relevant,
we encode the action history using a GRU: $\hidden{t} = \gru{\action{t}, \hidden{t-1}}$. Thus
the policy is $\policyOf{\obs{t}, \mem_{\ptr{t}}}$ or $\policyOf{\obs{t},
        \mem_{\ptr{t}}, \hidden{t}}$.

\subsection{Network architectures.}
\label{network-architectures}
Both $\edgeNet$, the network responsible for producing $\edgeChoice{t}$, and $\policy$ use a neural ``torso'' with shared weights, which is
specialized to handle the distinct observation spaces of each of our domains (\S\ref{starcraft} and \S\ref{minecraft}). The torso uses a convolutional neural network
for 3d components of the observation,
a lookup table of neural embeddings for integer components,
and a linear projection for all other components.
The torso concatenates the results of these operations,
applies a Rectified Nonlinear Unit \cite{nair2010rectified},
and passes the result
to the heads corresponding to $\policy$ and $\edgeNet$.
These heads are implemented using linear projections followed by a
softmax which transforms the outputs into a probability distribution
for sampling actions (in the case of $\policy$) and for choosing columns
of $\edges$ (in the case of $\phi$).

\subsection{Failure buffer.}
\label{failure-buffer}
During training the agent may learn a suboptimal policy which works most of the
time but fails for a small subset of instructions and episode starting
conditions. In order to encourage the agent to learn a policy that is robust to
these challenges, we modify the distribution of training episodes to increase the
frequency of difficult episodes. We accomplish this by saving the random seed
used to generate unsuccessful episodes to a ``failure buffer.'' We also maintain
a moving average of the agent's success-rate and, each episode, with probability
proportional to this success-rate, we sample the seed from the failure buffer to
retry a previously unsuccessful episode.

\subsection{Training details.}
\label{training-details}
We train the agent using Proximal Policy Optimization algorithm \cite{schulman2017proximal}, a
state-of-the-art on-policy RL algorithm. The learning objective is
\begin{align}
    L(\theta)   & =
    \mathbb{E}[\min(r_t(\theta)A_t, \clip\left(r_t(\theta), 1 - \epsilon, 1 +
        \epsilon\right)A_t] + \alpha\mathcal{H}
    \\
    r_t(\theta) & = \frac{\Pr(\action{t}, \ptrDelta{t},
        \gate{t}|\obs{t}, \memi{\ptr{t}}, \action{t}, \theta)}{\Pr(\action{t},
        \ptrDelta{t}, \gate{t}|\obs{t},\memi{\ptr{t}},
        \action{t}\theta_{\text{old}})}
    \label{r}
\end{align}
Here $A_t$ denotes the advantage on time step
$t$, $\theta$ denotes network parameters, $\theta_\text{old}$ represents the
pre-update parameters, $\alpha$ refers to an entropy coefficient, $\mathcal{H}$ refers to the entropy over the probability distribution in the numerator of (\ref{r}), and $\Pr$ refers to the joint probability of the action choice, pointer movement and gate, calculated as the product of their individual probabilities. For tuning hyper-parameters of all algorithms, we
searched for good common values for  hidden size, kernel size, and stride, for
both convolutions and for all hidden sizes used by neural networks. We also tuned
entropy coefficient values (used to encourage exploration), number of
distributions $\numEdges$ in $\edges$, and learning rate.

\section{Experiments}
In this section we present results from three generalization experiments in two
instruction-following domains, in which agents are trained on short instructions
and evaluated on longer instructions or instructions containing unseen
combinations of explicit control-flow blocks. The first domain is inspired by
the \textit{StarCraft} video game and is designed to impose challenges in learning
\textit{implicit} control flow. The second domain is inspired by \textit{Minecraft}
and is designed to impose challenges in learning \textit{explicit} control flow.
We compare our architecture to \del{two}\new{three} baselines, one using recurrence to maintain a
memory of progress through the instructions, \del{and the second }\new{another using} the OLSK
\cite{oh2017zero} architecture described above\new{, and a third using a modified version of OLSK with extended pointer-movement range}.
We also compare our architecture
to an ablation, \name{} which removes the \del{scan operation}\new{Scan mechanism} from \name{}-S. In all
results, error bands and bars indicate standard error across 4 distinct random
seeds.
To anticipate our main results: we find in both domains that \name{}-S
outperforms the \del{two }baselines as well as \name{} in generalization, especially
as instruction length increases in both domains.

\subsection{Baselines}
\label{baselines-description}


The \textbf{Unstructured Memory} \new{(UM)} baseline uses the recurrent state of a recurrent neural network
to track its progress through the instruction. Like
\name{}-S, this algorithm runs a bidirectional GRU length-wise across the
instructions. However, instead of retaining all of the outputs of the GRU and
maintaining a pointer, it feeds the concatenated last outputs of this
bidirectional GRU into a second GRU,
along with $\obs{t}$ and the embedded action $\action{t}$. \new{Note that the lengthwise GRU encoding of the instruction is necessary to facilitate generalization from shorter to longer instructions, which would not be possible if simpler methods like concatenation were used instead. Thus}
\del{Note that} the first GRU is responsible for encoding the variable-length instruction,
whereas the second is responsible for preserving state information across time steps.
The baseline must use the recurrent hidden state $\hidden{t}$ of the second GRU to
perform the functions that $\mem$ and $\ptr{t}$ perform in \name{}-S, tracking the
agent's progress through the instructions. The policy $\policy$
maps $\hidden{t}$ and the observation $\obs{t}$ to a distribution from which the
architectures samples the $\action{t}$.

The \textbf{OLSK} baseline reproduces the algorithm described in
OLSK \cite{oh2017zero}. Each time step, $\phi$ maps $\obs{t}$, $\memi{t}$, and a hidden state
$\hidden{t}$ to a
distribution over discrete pointer movements in $\set{-1, 0, +1}$ and a new hidden
state $\hidden{t+1}$ (per the hard-attention scheme described in \cite{oh2017zero}).
Thus,
OLSK can only move the pointer one step forward or one step
backward and so has to do this repeatedly without acting in the world to move the pointer many steps, in contrast to
\name{}-S, which can move the pointer between any two lines in the
instruction in one step.

\new{
    The \textbf{extended-range OLSK} (OLSK-E) extends the range of OLSK's pointer movement to
    facilitate pointer movements in $\set{-\instructionLength, \dots, +\instructionLength}$, where $\instructionLength$ is the length of the instruction.
    Note that OLSK does not use the bidirectional GRU to preprocess the instruction and it does not
    take advantage of the ``Scan'' mechanism described in \S\ref{scan}. Note that the absence of this
    preprocessing mechanism limits the ability of both OLSK and extended-range OLSK to
    operate in a context-aware manner.
}

The \textbf{\name{}} baseline is an ablation of \name{}-S. As the omission of
``-S'' suggests, \name{} ablates the Scan mechanism described in \S{\ref{scan}}
. Recall that \name{}-S passes the embedded instructions $\mem$
through a bidirectional GRU. \name{} retains only the final output of this GRU
and uses a single-layer neural network followed by a softmax layer to project
the  output to a distribution over forward/backward pointer movements, up to the
maximum length instruction to be evaluated. Note that while \name{} can in
principle move between any two lines, even in the longer evaluation
instructions, it will only have the opportunity during training to perform jumps
equal to or less than the size of the instruction.


\begin{figure}[t]
    \centering
    \begin{subfigure}[b]{.55\columnwidth}
        \resizebox{\columnwidth}{!}{

            \tikzstyle{background rectangle}=[fill=black]
            \tikzset{every node}\small\sffamily

            \begin{tikzpicture}[framed,scale=0.6]
                \node  at (2.5, 2.5) {\includegraphics[height=0.6cm,trim=6cm 0 6cm 0,clip]{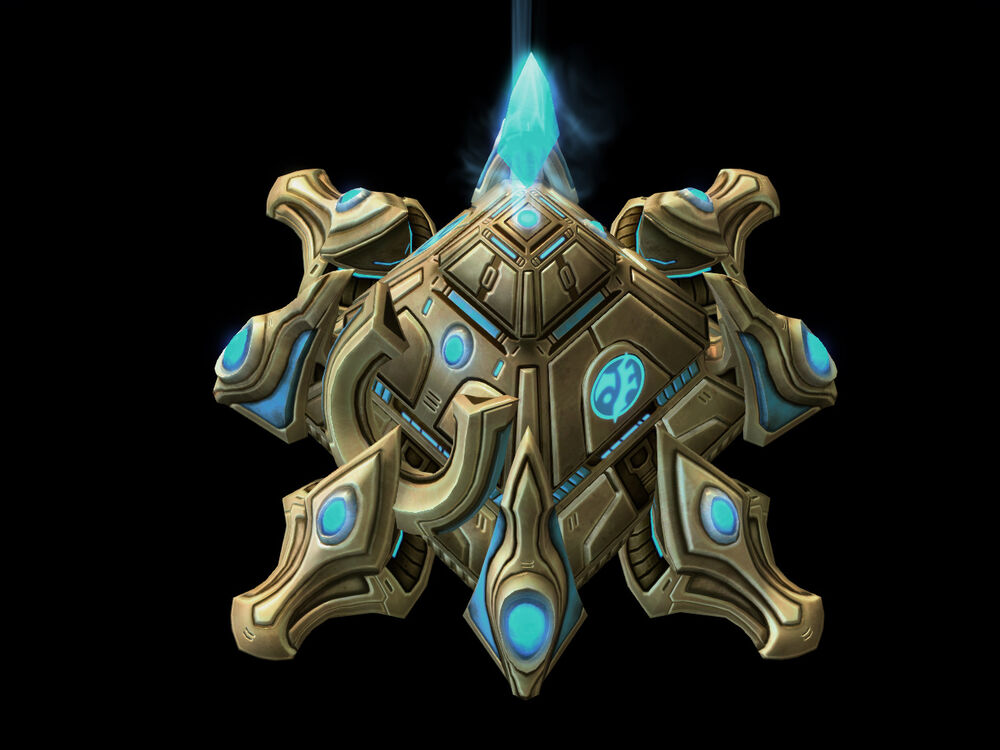}};
                \node  at (2.5, 3.5) {\includegraphics[height=0.6cm,trim=6cm 0 6cm 0,clip]{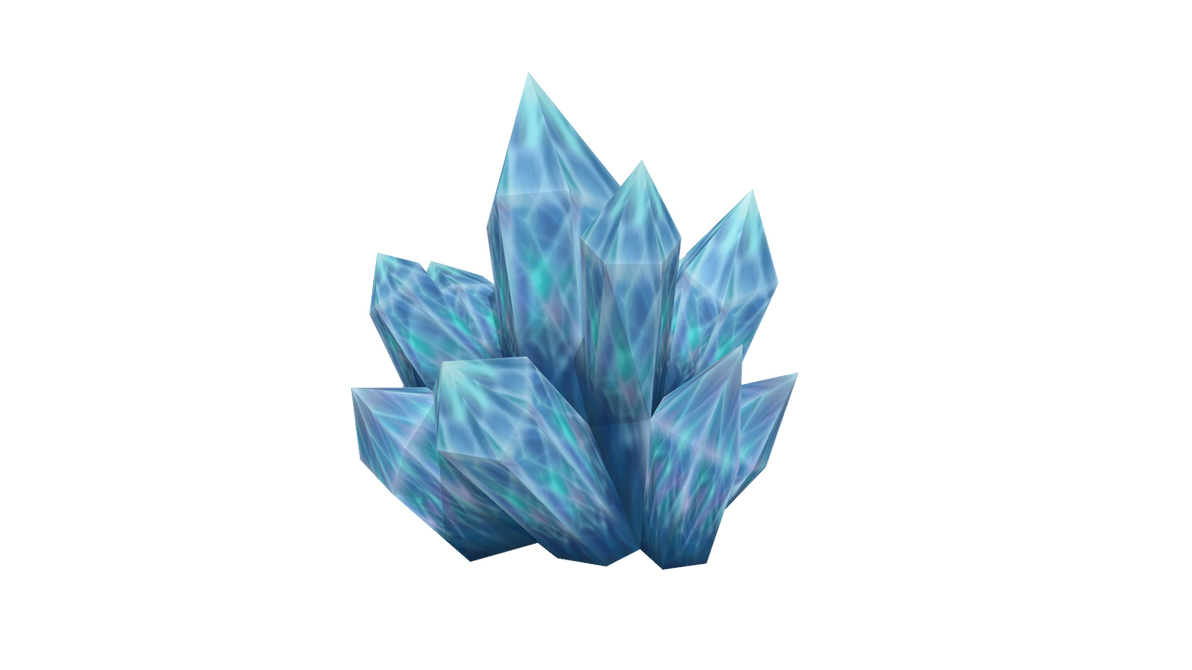}};
                \node  at (3.5, 2.5) {\includegraphics[height=0.6cm,trim=6cm 0 6cm 0,clip]{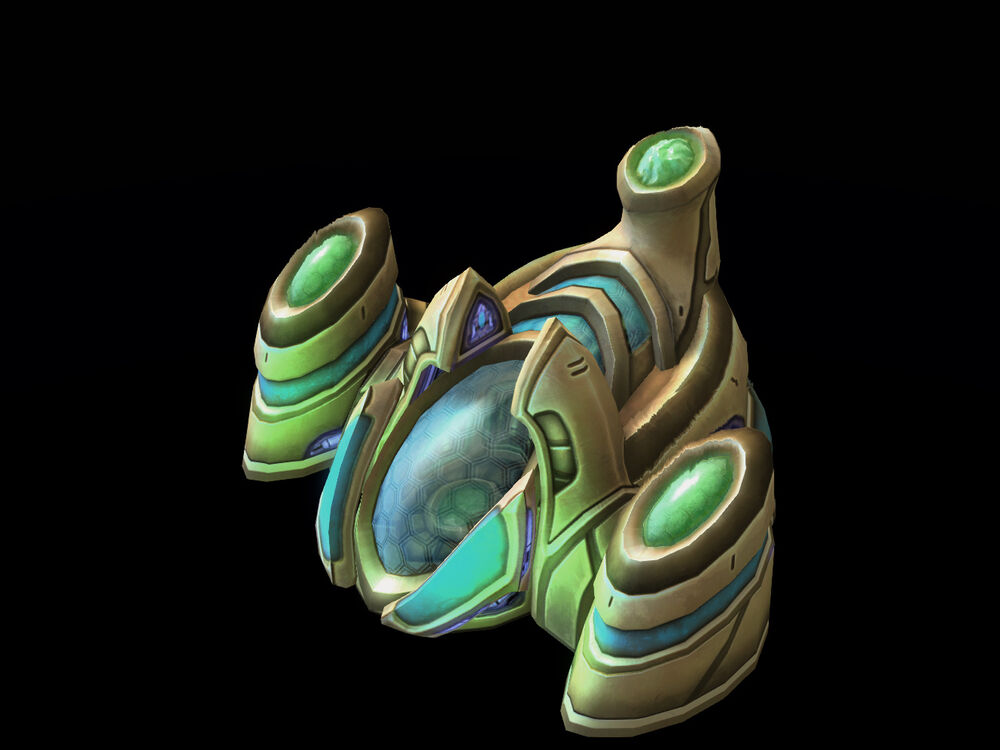}};
                \node  at (3.5, 3.5) {\includegraphics[height=0.6cm,trim=6cm 0 6cm 0,clip]{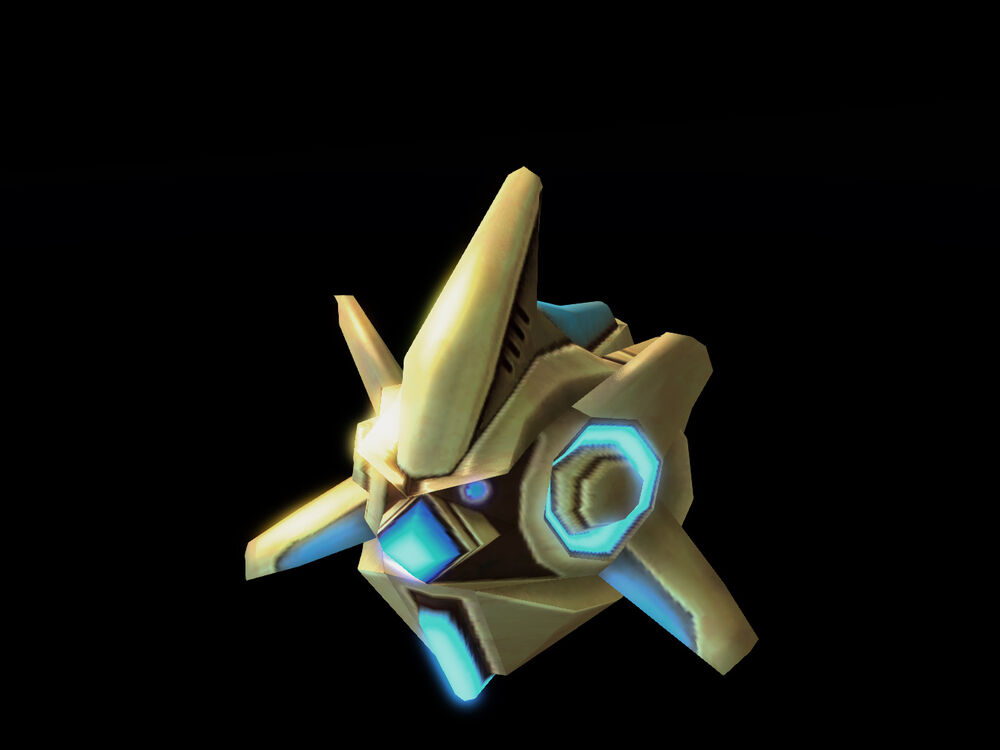}};
                \node  at (2.5, 0.5) {\includegraphics[height=0.6cm,trim=6cm 0 6cm 0,clip]{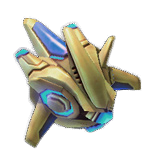}};
                \node  at (4.5, 0.5) {\includegraphics[height=0.6cm,trim=6cm 0 6cm 0,clip]{figures/probe}};
                \node  at (2.5, 5.5) {\includegraphics[height=0.6cm,trim=6cm 0 6cm 0,clip]{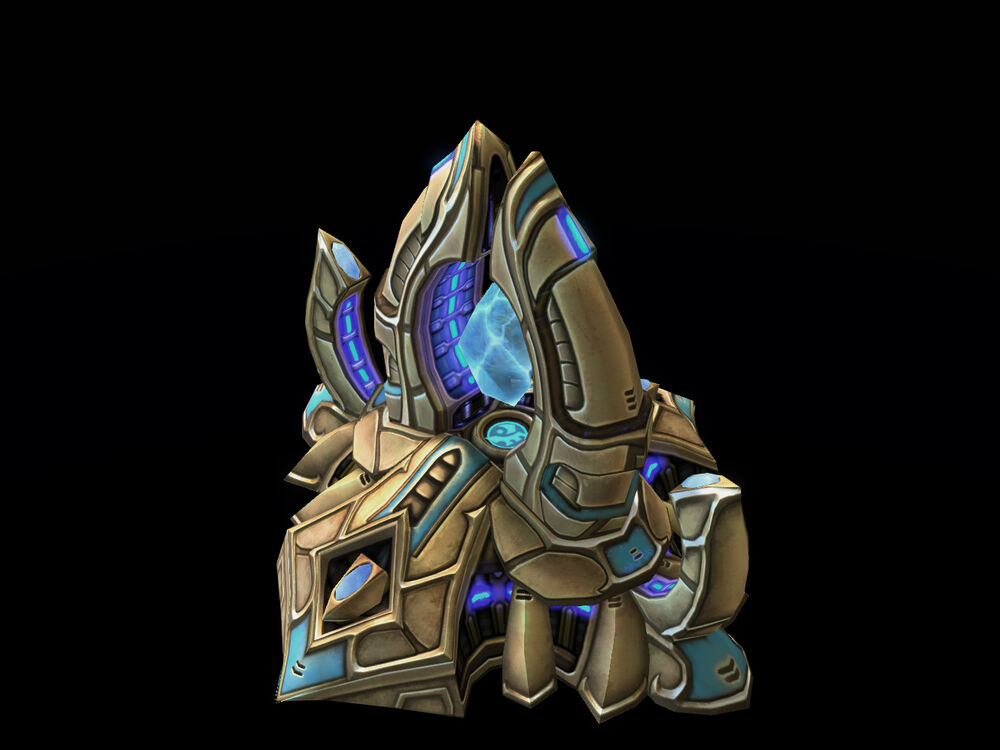}};
                \draw[color=gray,step=1] (0,0) grid (6,6);
                \node[text=white, font=\scriptsize]  (nex) at (1.5, 2.5) {nexus};
                \node[text=white, font=\scriptsize]  at (1.5, 0.4) {probe};
                \node[text=white, font=\tiny]        at (1.5, -0.2) {(builds buildings)};
                \node[text=white, font=\scriptsize]  (ass) at  (5.0, 2.5) {assimilator};
                \node[text=white, font=\scriptsize]  (gat) at (3.8, 5.5) {gateway};
            \end{tikzpicture}
        }
    \end{subfigure}
    \hfill
    \begin{minipage}[b]{0.4\columnwidth}
        \begin{subfigure}[b]{\columnwidth}
            {\scriptsize

                \Tree [.{Nexus (trains Adept)} {\ldots} {\ldots} {\ldots} ]

                \Tree [.{Assimilator} [.{Gateway (trains Colossus)} ] {\ldots} ]
            }
        \end{subfigure}
        \\
        \begin{subfigure}[b]{\columnwidth}
            \begin{lstlisting}[style=base]
Build Nexus
Train Adept
Build Assimilator
Build Gateway
Train Colossus
\end{lstlisting}
            \label{starcraft:instruction}
        \end{subfigure}
    \end{minipage}
    \caption{Example of StarCraft-inspired environment and instructions. The
        instructions indicate that Nexus is the building which trains the Adept
        unit, the Assimilator is a prerequisite for the
        Gateway, and the Gateway trains the Colossus. The randomly generated
        build-tree (upper right) is the basis for the instructions, but the
        agent only sees the instructions, not the tree, and the instructions
        only contain information relevant to the production of the required units
        (the Adept and the Colossus, in this example).
    }
    \label{fig:starcraftexample}
\end{figure}

\begin{figure*}[ht]
    \begin{subfigure}[b]{.71\columnwidth}
        \includegraphics[width=0.9\columnwidth]{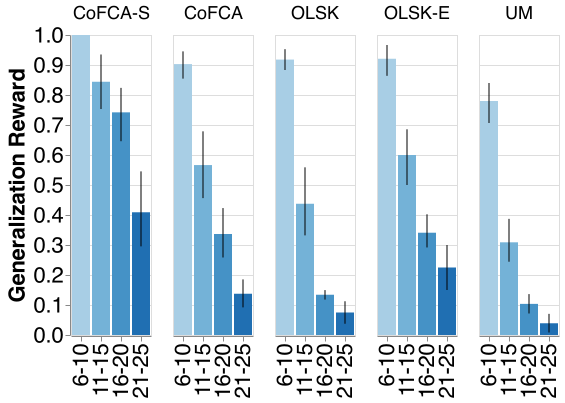}
    \end{subfigure}%
    \hfill
    \begin{subfigure}[b]{.67\columnwidth}
        \includegraphics[width=0.9\columnwidth]{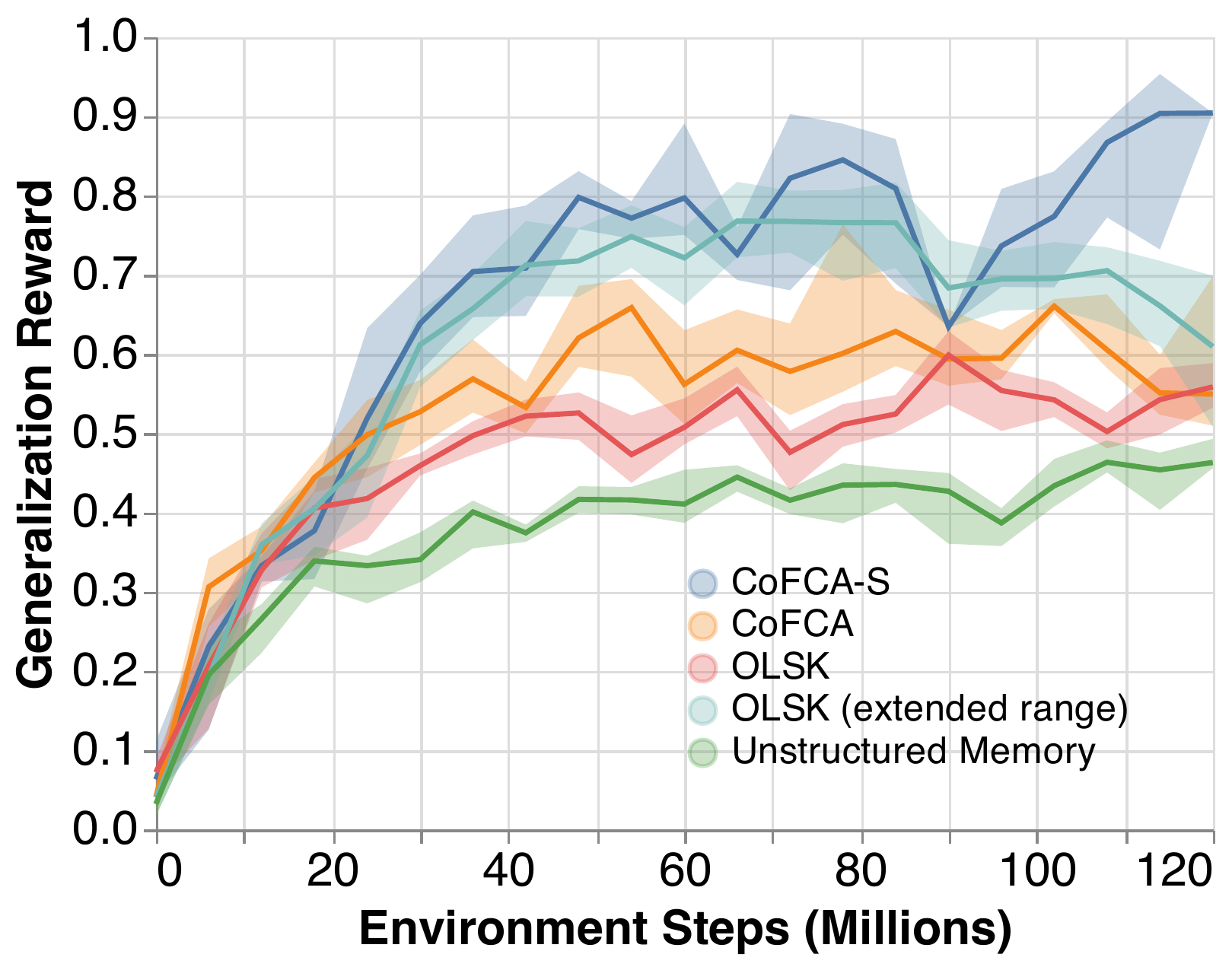}
    \end{subfigure}%
    \hfill
    \begin{subfigure}[b]{.66\columnwidth}
        \includegraphics[width=0.9\columnwidth]{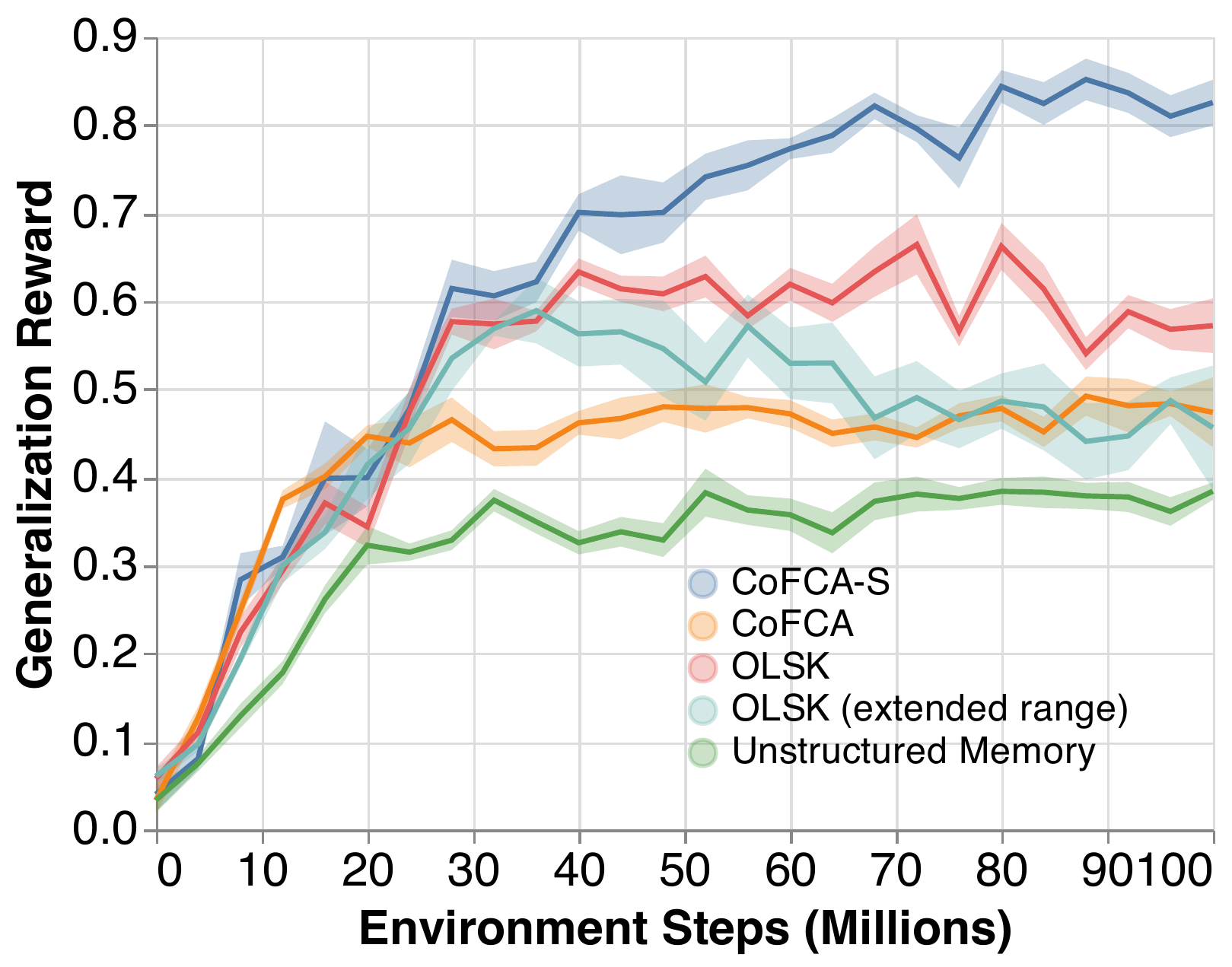}
    \end{subfigure}%
    \caption{
        Results for \new{experiments in the StarCraft-inspired domain (\S \ref{starcraft})}\del{Experiment 1: Generalization of learned implicit control flow  in the StarCraft-inspired domain}.
        Y-axes represents cumulative reward on evaluation episodes.
        (Left)
        Generalization to longer instructions (\S \ref{exp:starcraft-longer}), binned by instruction length\new{, aggregated from the last 5 million steps}.
        (Middle)
        Learning curves for generalization to longer instruction lengths.
        \new{(Right)
            Learning curves for generalization to longer instruction lengths and deeper build trees (\S \ref{exp:starcraft-deeper}).}
    }
    \label{starcraft-results}
\end{figure*}

\subsection{StarCraft-inspired domain: implicit control flow}
\label{starcraft}
StarCraft is a real-time strategy game in which players construct
buildings to produce units and research technologies, and use those units to
destroy an opponent. In our StarCraft-inspired domain the agent assumes the role of a
support player that receives an order for units from an allied player; these orders are the instructions given to the agent. To satisfy the order for a type of unit, the agent must construct the buildings that produce that
type of unit and then train the units.
The agent cannot construct a
building until a prerequisite building has been constructed.
Each building depends on
at most one building, though each building may serve as a prerequisite for
several others, forming a \textit{build tree} of dependencies. Figure~\ref{fig:starcraftexample} provides an example.

Two aspects of the environment dynamics interact to introduce implicit control flow learning challenges.
The first is that enemy attacks on buildings and ambushes on units happen stochastically, requiring buildings to be reconstructed and units to be retrained. The second is that the type of buildings that produce each unit type as well as the build-tree dependencies are randomly generated for each episode, and this information is encoded in the instructions, requiring the agent to
learn to extract these relationships from the instructions instead of  learning them from experience.  Also, when the enemy ambushes units or attacks buildings, the agent must learn to respond to these events and re-execute just those parts of the instructions whose effects were undone.


Episodes play out in a $6\times 6$ gridworld. The agent begins each episode with
three Probes (units for constructing buildings) and a random
endowment of pre-constructed buildings. If these buildings correspond to
buildings required by the instruction, the agent may opportunistically skip the
corresponding lines of instruction. The agent's
observation includes a top-down view of the gridworld with channels devoted to
buildings, units, and terrain.

\paragraph{Instructions.}
Each line in the instruction
is encoded by an integer indicating either a building type or a unit type. A building followed by a
building indicates that the first is a prerequisite of the second. A
building followed by a unit indicates that the building must be used to build
the unit.

\paragraph{Attacks and ambushes.}
Enemy attacks and ambushes occur each time-step with 10\% probability. An attack
wipes out a random subset of the existing buildings. An
ambush destroys one of the units the agent has produced. If the instructions indicated that the destroyed unit
was required, the agent must produce the unit again.
The agent may also need to rebuild the buildings required to construct
that unit if the building was previously destroyed in an enemy attack,
requiring the agent to 
consult earlier sections of the instructions.
Messages to the agent notifying it about
unit ambushes assume the form of a single integer, which identifies the type of unit ambushed.

\paragraph{Reward and termination.}
The agent receives a reward of 0 on each timestep and 1 for completing the
instructions (producing all required units). The episode terminates when the
instructions are completed or when a time limit runs out. The time limit is
$30 \times$ the length of the instructions.

\paragraph{Modifications to the \del{four }architectures to handle the large action space\del{ and partial observability}.}
\del{We made modifications to the architectures to handle two issues arising in this domain.
    The first  issue is the large, structured action space.}
There are 1512 build commands (3 Probes
$\times$ 36 coordinates $\times$ 14 buildings) and 108 go-to commands (3 Probes
$\times$ 36 coordinates). Buildings are chosen by coordinate and therefore there
are as many as 576 train-unit commands (36 coordinates $\times$ 16 unit types).
To handle this action space we adopt an autoregressive policy
\cite{metz2017discrete}, which allows the agent to select a variable-length
\textit{sequence} of actions between environment interactions. The agent
first chooses a Probe or a coordinate. Choosing a coordinate containing a
building selects that building and allows the agent to produce any unit
that the building is capable of producing. Choosing a Probe allows the agent to
construct buildings using that Probe or to move the Probe to a
coordinate. Constructing a building requires choosing a building then an
unoccupied coordinate. Invalid choices (e.g. telling a Probe to construct a
building for which the prerequisite has not been constructed) result in a no-op.
This dependence on the history of actions motivates the addition of the action GRU described
in \S\ref{action-sampling}.

\del{The second issue is that the agent can influence the actions of the Probe for several
    time-steps, introducing partial observability; an
    optimal agent will need to remember previously given commands. To handle this,
    we pass the current action through a recurrent GRU. {In COFCA-S, we also feed an encoding of
            $\obs{t}$ through the bidirectional GRU along with $\mem$ to produce
            $\bigruOutput$. This allows the agent to condition pointer movements on the
            changing conditions in the environment.}}


\subsubsection{Generalization to longer instructions under implicit control flow demands} 
\label{exp:starcraft-longer}

\paragraph{Training and evaluation.}
The aim of this experiment is to test the ability of the \del{four }agents \del{(\name{}-S,
    \name{}, OLSK, Unstructured Memory) }to learn implicit control flow strategies
from short instructions and generalize to longer ones. All \del{four }agents  were
trained in the StarCraft-inspired domain on randomly sampled instructions of
length 1 to 5. 
Every million frames, we evaluated agent performance on 150 complete
episodes with instructions of length 6 to 25. In this way we
tested the ability of the agent to
learn strategies for re-executing parts of the instructions affected by disruptive events, in a way that generalizes to longer instructions. 
\paragraph{Results.}
Figure \ref{starcraft-results} shows the performance of all \del{four}\new{five} architectures
on the evaluation instructions, with performance binned by instructions length.
\name{}-S outperforms the baselines on all instructions lengths. We conjecture
that Unstructured Memory has difficulty tracking its place in non-sequential
control flow, and  OSLK has difficulty performing large pointer movements
through a sequence of single-steps.
We conjecture \name{}-S performs best because it can learn a logic of pointer
movement that is \textit{independent} of the size of the movement. 
For example, it might learn to scan backward to the first of a series of buildings
destroyed by the enemy. Such logic can transfer from the short training instructions to longer evaluation instructions.
\new{
    \subsubsection{Generalization to longer instructions and deeper build trees} }
\label{exp:starcraft-deeper}
\new{\paragraph{Training and Evaluation.} This experiment reproduces the setup of the previous (\S \ref{exp:starcraft-longer}) but
    during training, it restricts the depth of build-trees to three. Though the agent is free to perform larger jumps, the agent need
    only perform jumps of size three or less to complete these tasks.
    Meanwhile, we evaluate the agents on trees with unrestricted depth (up to 16, if the tree forms a chain). Thus these experiments
    increases the pressure on agents to perform longer jumps during evaluation than during training.
}
\new{
    \paragraph{Results.}
    This experiment highlights the advantages of the
    Scan mechanism which is designed to facilitate generalization from shorter jumps during training to longer jumps in evaluation.
    OLSK's second-best performance makes sense given that it does not perform jumps at all, and is therefore less susceptible to
    overfitting on shorter jumps learned during training.
}

\subsection{Minecraft-inspired domain: explicit control flow}
\label{minecraft}
In the prior StarCraft-inspired domain the instructions themselves imposed a simple linear execution order, and the interesting non-linear control flow challenges arose from stochastic environment dynamics. In this section we explore non-linear control flow that is imposed by the instructions themselves through explicit control flow elements.

Figure~\ref{fig:example} provides an example of our domain (inspired by
\citet{sun2020program} and the Minecraft video game).
The agent navigates a  $6\times6$ gridworld in which resources, terrain,
and merchants spawn randomly each episode.
In each episode the agent is given  new instructions containing
subtasks interspersed with control-flow statements. The agent's goal is to
perform these subtasks in the order specified by the control flow.
The agent's observation includes a
top-down view of this grid with channels encoding the presence of
objects in the gridworld. The resources are \emph{gold}, \emph{iron}, and
\emph{wood}. The terrain includes impassible \emph{walls} and
\emph{water} which can be bridged only if the agent possesses wood in its inventory.

\begin{figure}[t]
    \centering
    \begin{subfigure}[b]{.45\columnwidth}
        \resizebox{\columnwidth}{!}{
            \begin{tikzpicture}[scale=0.6]
                \draw[color=gray] (0,0) grid (6,6);
                \node (agent)  at (5.5, 5) {\pgfbox[center,bottom]{\pgfuseimage{agent}}};
                \node (bridg) at (3.5, 1) {\pgfbox[center,bottom]{\pgfuseimage{bridge}}};
                \node (gold) at (0.5, 2) {\pgfbox[center,bottom]{\pgfuseimage{gold}}};
                \node (iron1) at (4.5, 0) {\pgfbox[center,bottom]{\pgfuseimage{iron}}};
                \node (iron2) at (0.5, 0) {\pgfbox[center,bottom]{\pgfuseimage{iron}}};
                \node (iron3) at (1.5, 3) {\pgfbox[center,bottom]{\pgfuseimage{iron}}};
                \node (merc1) at (2.5, 5) {\pgfbox[center,bottom]{\pgfuseimage{merchant}}};
                \node (merc2) at (1.5, 0) {\pgfbox[center,bottom]{\pgfuseimage{merchant}}};
                \node (wate1) at (3.5, 0) {\pgfbox[center,bottom]{\pgfuseimage{water}}};
                \node (wate2) at (3.5, 2) {\pgfbox[center,bottom]{\pgfuseimage{water}}};
                \node (wate3) at (3.5, 3) {\pgfbox[center,bottom]{\pgfuseimage{water}}};
                \node (wate4) at (3.5, 4) {\pgfbox[center,bottom]{\pgfuseimage{water}}};
                \node (wate5) at (3.5, 5) {\pgfbox[center,bottom]{\pgfuseimage{water}}};
                \node (wood1) at (2.5, 3) {\pgfbox[center,bottom]{\pgfuseimage{wood}}};
                \node (wood2) at (2.5, 4) {\pgfbox[center,bottom]{\pgfuseimage{wood}}};
                \node (wood3) at (5.5, 2) {\pgfbox[center,bottom]{\pgfuseimage{wood}}};
                \node (wood4) at (5.5, 0) {\pgfbox[center,bottom]{\pgfuseimage{wood}}};
                \draw [->, thick, orange] (5.5, 5) -- (5.5,0.5) -- (5, 0.5);
                \draw [->, thick, orange] (4.5, 1) -- (4.5,1.5) -- (2.5,1.5) -- (2.5,1.5) -- (2.5,3.5) -- (2,3.5);
                \draw [->, thick, red] (1.5, 3) -- (1.5,0.5) -- (1,0.5);
                \draw [->, thick, violet] (0.5,1) -- (0.5,2);
                \draw [->, thick, violet] (0.5,3) -- (0.5,5.5) -- (2,5.5);
            \end{tikzpicture}
        }
    \end{subfigure}
    \hfill
    \begin{subfigure}[b]{.49\columnwidth}
        \begin{lstlisting}[style=base]
while more iron than gold
   @mine iron@
endwhile
if more merchants than iron
   ^inspect iron^
   &sell gold&
else
  mine wood
endif
\end{lstlisting}
        \label{minecraft:instruction}
    \end{subfigure}
    \caption{Example of instructions and environment state in the Minecraft-inspired domain. Correct execution begins
        at the first line with evaluation of the \emph{while} condition, which
        evaluates to true because there is more \emph{iron} than \emph{gold}. To
        perform the \emph{mine iron} subtask, the agent must navigate to an iron
        resource (the orange arrow) and perform a \emph{mine} action. This
        removes the iron resource from the environment but the \emph{while}
        condition still evaluates to true because there is still more iron than
        gold. After mining a second iron resource (second orange arrow), the
        numbers of iron and gold resources are equal and the \emph{while}
        condition evaluates to false. At this point, the number of
        \emph{merchants} in the environment (depicted as llamas) exceeds the
        number of iron resources and the \emph{if} condition evaluates to true.
        The agent must then execute \emph{inspect iron} and \emph{sell gold} but
        must skip  the \emph{mine wood} subtask. To execute \emph{sell gold} the
        agent must navigate to a gold resource and mine it (first violet arrow),
        navigate to a merchant, and then sell the gold (second violet arrow),
        terminating the episode with a reward of 1.}
    \label{fig:example}
\end{figure}

\paragraph{Instructions.}
At the start of each episode, the agent receives instructions
comprised of a list of lines, each corresponding 
to a subtask or a control-flow keyword: \emph{if}, \emph{else},
\emph{endif}, \emph{while}, or \emph{endwhile}. These have the procedural
semantics familiar from programming languages.  (The indentation in
Figure \ref{fig:example} is a visual aid not available to the agent.)
Each line in the instructions consists of three symbols:
the first encodes the control-flow keyword or indicates that the line is a subtask;
the second encodes an action; the third encodes either the target
resource for the action or the condition for an \emph{if} or \emph{while}
line that requires resource comparisons; e.g. \emph{more iron than gold} is true if the amount of iron in the environment exceeds the amount of gold.
While the agent observes the
entire instructions on each time-step, nothing in the observation indicates the
agent's {progress} through the instructions.

\paragraph{Actions and inventory.}
The agent acts by issuing verb-noun commands to a worker. The possible verbs are
\emph{mine}, \emph{sell}, and \emph{inspect}, while the possible nouns
correspond to resources: \emph{iron}, \emph{gold}, and \emph{wood}. The
action-space of the agent is the cross-product of nouns and verbs. The movements
of the worker in response to these commands are determined by a pre-trained RL
agent. In response to a \emph{mine resource} command, the worker navigates to
the resource and executes a \emph{mine} action, removing it from the map and
adding it to the inventory. For a \emph{sell resource} command, the  worker
executes a \emph{sell} interaction on a merchant grid, which decrements the
resource inventory. If the worker does not already possess the resource,
it must first collect it using the \emph{mine} action. To complete an
\emph{inspect resource} command, the worker must execute an \emph{inspect} action
on a resource grid (this does not \del{mutate}\new{change or remove the resource}\del{the environment}).
If the agent's inventory
contains wood, the agent may move into a water grid, decrementing the wood in its
inventory and removing the water grid (building a bridge).
The agent observes the inventory as a list of integers encoding the quantity of each resource type.

\paragraph{Reward and termination.}
The agent receives a reward of 1 upon completion of all subtasks in the
instructions, in the order specified by control flow. All other time steps
provide a reward of 0. An episode terminates when the instructions are completed or if the worker performs a \emph{mine}
or \emph{sell} action out of the order specified by the instructions, or if the
agent exceeds a time-limit equal to $30 \times$ 
the length of the instructions.

\begin{figure*}[ht]
    \begin{subfigure}[b]{.78\columnwidth}
        \includegraphics[width=\columnwidth]{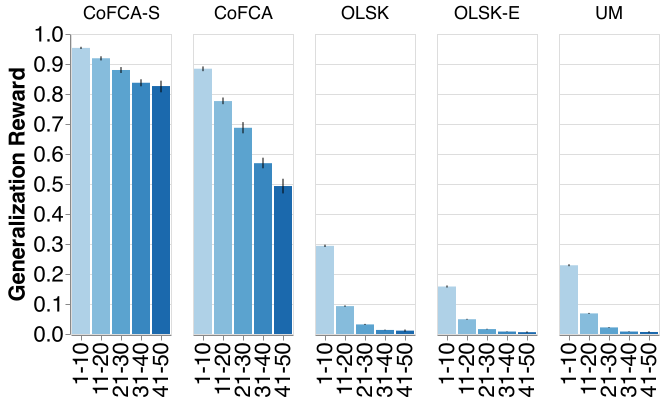}
    \end{subfigure}%
    \hfill
    \begin{subfigure}[b]{.57\columnwidth}
        \includegraphics[width=\columnwidth]{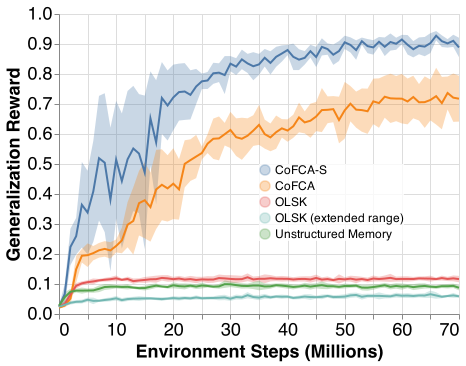}
    \end{subfigure}%
    \hfill
    \begin{subfigure}[b]{.57\columnwidth}
        \includegraphics[width=\columnwidth]{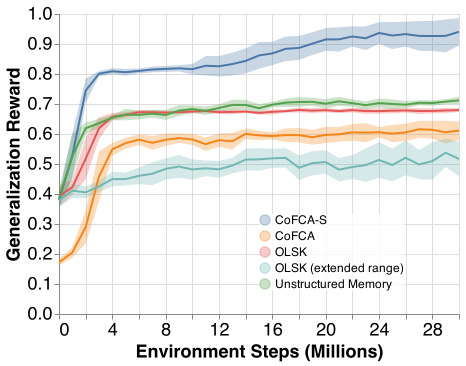}
    \end{subfigure}%
    \label{subfig:compound}
    \caption{
        Results for experiments in the Minecraft-inspired  domain.
        Y-axes represents cumulative reward on evaluation episodes.
        (Left)
        Generalization to longer \textbf{instructions length}, binned by instruction (\S\ref{minecraft-main-results})
        length
        (Middle)
        Learning curves for generalization to \textbf{longer instructions} (\S\ref{minecraft-main-results}).
        (Right)
        Learning curves for generalization to instructions with \textbf{novel compositions of control-flow} (\S\ref{minecraft-compound-results}).
        }
    \label{fig:minecraft-length}
\end{figure*}%


\subsubsection{Generalization to longer instructions with explicit control flow}
\label{minecraft-main-results}

\paragraph{Training and evaluation.}
The  aim of this experiment is to test the \del{four }agents' ability to learn how to
follow explicit control flow from short instructions and generalize to  longer
instructions. We trained the \del{four }architectures on instructions of length 1 to
10 in our Minecraft-inspired domain, randomly sampled from a generative grammar
(see Appendix). Every 100 gradient steps, we evaluated the
performance of the agent for 500 time steps on instructions of length 11 to 50,
also sampled from the same grammar.

\paragraph{Results.}
\label{exp:instruction-length}

Figure~\ref{fig:minecraft-length} (left and middle) provides the evaluation results. \name{}-S and
\name{} outperform the \del{two}\new{three} baselines, \del{both}\new{all} of which fail to generalize
entirely.
In the case of OLSK this is again likely because the pointer cannot make
individual movements greater than $\pm1$, and instead must make a series of
these small movements. \new{Conversely OLSK-E is capable of making larger movements,
    but lacks the bidirectional GRU preprocessing step that would give it access to
    information about lines surrounding the $\ptr{t}^\text{th}$ and enable it to make these larger
    movements judiciously. Instead, OLSK-E must derive this kind of information
    from its memory of previously visited lines.
    Inspecting the trajectories of OLSK-E, we observe that it ultimately gives up on
    learning to manipulate the pointer, ignoring the instruction and instead learning a prior
    over the subtasks.
} There is also a smaller but still significant gap
between the generalization performance of \name{}-S and \name{}.
In Figure~\ref{fig:minecraft-length} (left) we break down the performance
of all the agents after training for 70 million time steps as a function of the
length of the instructions. As expected each agent's performance degrades with
increasing length of instructions, though for both \name{} and \name{}-S the
drop in generalization performance from the shortest generalization lengths
$10$-$20$ to the longest generalization lengths of $40$-$50$ is much smaller
than for the baselines. 

\subsubsection{Generalization to novel compositions of explicit control flow}
\label{minecraft-compound-results}

\paragraph{Training and evaluation.}
To test generalization to novel control flow compositions, we trained the \del{four }agents in the MineCraft-inspired domain on instructions of length up to $10$, excluding those that contain more than one type of control flow
(though possibly more than one instance of the same type of control flow).
Every 100 gradient steps, we evaluated the performance of the agent for 500 time steps
on instructions of length up to $10$,  but \textit{only}
those that contained at least two different types of control flow in them. (The maximum instruction length for testing and training was the same so as not to confound the effects with length.)

\paragraph{Results.}
All architectures do better in this experiment because the instructions are shorter, but again \name{}-S dominates.  
We attribute the weak performance of \name{} to its
dependence on the bidirectional GRU without the benefit of the Scan mechanism. We conjecture that a recurrent
network that has only ever trained on instructions of one type is likely to
generate noisy outputs when first encountering instructions with multiple types of control flows
--- e.g., seeing an \emph{if} line following a \emph{while} line.
While the \name{}-S architecture also depends on recurrence,
the Scan mechanism ensures that the potentially noisy outputs of
the GRU outside any given control-flow block are mostly ignored.

\section{Conclusion}
This work contributes a neural network architecture with a novel attentional mechanism that moves pointers based on where the pointer should move next rather than how much it should move and also processes instructions using the current pointer position. As a result, our architecture allows RL agents
to learn to follow instructions with implicit and explicit control-flow as well as to generalize better to longer and novel instructions. Our empirical work
demonstrated this benefit in two domains.

\section*{Acknowledgements}
This work was made possible by the support of the Lifelong Learning Machines (L2M) grant from the
Defense Advanced Research Projects Agency.
Any opinions, findings, conclusions, or recommendations expressed here are those of the authors and do not necessarily reflect the views of the sponsors.
\bibliography{ms}

\begin{thebibliography}{19}
\providecommand{\natexlab}[1]{#1}
\providecommand{\url}[1]{\texttt{#1}}
\expandafter\ifx\csname urlstyle\endcsname\relax
  \providecommand{\doi}[1]{doi: #1}\else
  \providecommand{\doi}{doi: \begingroup \urlstyle{rm}\Url}\fi

\bibitem[Bahdanau et~al.(2018)Bahdanau, Hill, Leike, Hughes, Hosseini, Kohli,
  and Grefenstette]{bahdanau2018learning}
Bahdanau, D., Hill, F., Leike, J., Hughes, E., Hosseini, A., Kohli, P., and
  Grefenstette, E.
\newblock Learning to understand goal specifications by modelling reward.
\newblock \emph{arXiv preprint arXiv:1806.01946}, 2018.

\bibitem[Bieber et~al.(2020)Bieber, Sutton, Larochelle, and
  Tarlow]{bieber2020learning}
Bieber, D., Sutton, C., Larochelle, H., and Tarlow, D.
\newblock Learning to execute programs with instruction pointer attention graph
  neural networks.
\newblock \emph{arXiv preprint arXiv:2010.12621}, 2020.

\bibitem[Bo{\v{s}}njak et~al.(2017)Bo{\v{s}}njak, Rockt{\"a}schel, Naradowsky,
  and Riedel]{bovsnjak2017programming}
Bo{\v{s}}njak, M., Rockt{\"a}schel, T., Naradowsky, J., and Riedel, S.
\newblock Programming with a differentiable forth interpreter.
\newblock In \emph{Proceedings of the 34th International Conference on Machine
  Learning-Volume 70}, pp.\  547--556. JMLR. org, 2017.

\bibitem[Chaplot et~al.(2018)Chaplot, Sathyendra, Pasumarthi, Rajagopal, and
  Salakhutdinov]{chaplot2018gated}
Chaplot, D.~S., Sathyendra, K.~M., Pasumarthi, R.~K., Rajagopal, D., and
  Salakhutdinov, R.
\newblock Gated-attention architectures for task-oriented language grounding.
\newblock In \emph{Proceedings of the AAAI Conference on Artificial
  Intelligence}, volume~32, 2018.

\bibitem[Chung et~al.(2014)Chung, Gulcehre, Cho, and
  Bengio]{chung2014empirical}
Chung, J., Gulcehre, C., Cho, K., and Bengio, Y.
\newblock Empirical evaluation of gated recurrent neural networks on sequence
  modeling.
\newblock \emph{arXiv preprint arXiv:1412.3555}, 2014.

\bibitem[Fried et~al.(2018)Fried, Hu, Cirik, Rohrbach, Andreas, Morency,
  Berg-Kirkpatrick, Saenko, Klein, and Darrell]{fried2018speaker}
Fried, D., Hu, R., Cirik, V., Rohrbach, A., Andreas, J., Morency, L.-P.,
  Berg-Kirkpatrick, T., Saenko, K., Klein, D., and Darrell, T.
\newblock Speaker-follower models for vision-and-language navigation.
\newblock \emph{arXiv preprint arXiv:1806.02724}, 2018.

\bibitem[Graves et~al.(2013)Graves, Jaitly, and Mohamed]{graves2013hybrid}
Graves, A., Jaitly, N., and Mohamed, A.-r.
\newblock Hybrid speech recognition with deep bidirectional lstm.
\newblock In \emph{2013 IEEE workshop on automatic speech recognition and
  understanding}, pp.\  273--278. IEEE, 2013.

\bibitem[Graves et~al.(2014)Graves, Wayne, and Danihelka]{graves2014neural}
Graves, A., Wayne, G., and Danihelka, I.
\newblock Neural turing machines.
\newblock \emph{arXiv preprint arXiv:1410.5401}, 2014.

\bibitem[Hill et~al.(2020)Hill, Mokra, Wong, and Harley]{hill2020human}
Hill, F., Mokra, S., Wong, N., and Harley, T.
\newblock Human instruction-following with deep reinforcement learning via
  transfer-learning from text.
\newblock \emph{arXiv preprint arXiv:2005.09382}, 2020.

\bibitem[Metz et~al.(2017)Metz, Ibarz, Jaitly, and Davidson]{metz2017discrete}
Metz, L., Ibarz, J., Jaitly, N., and Davidson, J.
\newblock Discrete sequential prediction of continuous actions for deep rl.
\newblock \emph{arXiv preprint arXiv:1705.05035}, 2017.

\bibitem[Misra et~al.(2018)Misra, Bennett, Blukis, Niklasson, Shatkhin, and
  Artzi]{misra2018mapping}
Misra, D., Bennett, A., Blukis, V., Niklasson, E., Shatkhin, M., and Artzi, Y.
\newblock Mapping instructions to actions in 3d environments with visual goal
  prediction.
\newblock \emph{arXiv preprint arXiv:1809.00786}, 2018.

\bibitem[Nair \& Hinton(2010)Nair and Hinton]{nair2010rectified}
Nair, V. and Hinton, G.~E.
\newblock Rectified linear units improve restricted boltzmann machines.
\newblock In \emph{Proceedings of the 27th international conference on machine
  learning (ICML-10)}, pp.\  807--814, 2010.

\bibitem[Oh et~al.(2017)Oh, Singh, Lee, and Kohli]{oh2017zero}
Oh, J., Singh, S., Lee, H., and Kohli, P.
\newblock Zero-shot task generalization with multi-task deep reinforcement
  learning.
\newblock In \emph{Proceedings of the 34th International Conference on Machine
  Learning-Volume 70}, pp.\  2661--2670. JMLR. org, 2017.

\bibitem[Reed \& De~Freitas(2015)Reed and De~Freitas]{reed2015neural}
Reed, S. and De~Freitas, N.
\newblock Neural programmer-interpreters.
\newblock \emph{arXiv preprint arXiv:1511.06279}, 2015.

\bibitem[Schulman et~al.(2017)Schulman, Wolski, Dhariwal, Radford, and
  Klimov]{schulman2017proximal}
Schulman, J., Wolski, F., Dhariwal, P., Radford, A., and Klimov, O.
\newblock Proximal policy optimization algorithms.
\newblock \emph{arXiv preprint arXiv:1707.06347}, 2017.

\bibitem[Sohn et~al.(2018)Sohn, Oh, and Lee]{sohn2018hierarchical}
Sohn, S., Oh, J., and Lee, H.
\newblock Hierarchical reinforcement learning for zero-shot generalization with
  subtask dependencies.
\newblock In \emph{Advances in Neural Information Processing Systems}, pp.\
  7156--7166, 2018.

\bibitem[Sun et~al.(2020)Sun, Wu, and Lim]{sun2020program}
Sun, S.-H., Wu, T.-L., and Lim, J.~J.
\newblock Program guided agent.
\newblock In \emph{International Conference on Learning Representations}, 2020.
\newblock URL \url{https://openreview.net/forum?id=BkxUvnEYDH}.

\bibitem[Yu et~al.(2018{\natexlab{a}})Yu, Lian, Zhang, and Xu]{yu2018guided}
Yu, H., Lian, X., Zhang, H., and Xu, W.
\newblock Guided feature transformation (gft): A neural language grounding
  module for embodied agents.
\newblock In \emph{Conference on Robot Learning}, pp.\  81--98. PMLR,
  2018{\natexlab{a}}.

\bibitem[Yu et~al.(2018{\natexlab{b}})Yu, Zhang, and Xu]{yu2018interactive}
Yu, H., Zhang, H., and Xu, W.
\newblock Interactive grounded language acquisition and generalization in a 2d
  world.
\newblock \emph{arXiv preprint arXiv:1802.01433}, 2018{\natexlab{b}}.

\end{thebibliography}
\bibliographystyle{icml2021}

\normalem

\twocolumn[
    \icmltitle{Appendix}
]
\renewcommand\thesection{\Alph{section}}
\setcounter{section}{0}
\section{StarCraft environment details}
\subsection{Instruction grammar}

\begin{grammar}

    <task> ::= \{ `,' <line-type> \}

    <line-type> ::= <building> | <unit>
\end{grammar}
\subsection{Generation of instructions}
\label{starcraft:instructions}
We initially generate the build-tree as a random directed acyclic graph.
Next we randomize the building production capabilities by assigning a random
building to each unit (that building being the one capable of producing that unit).
We generate instructions, unit by unit, randomly selecting each unit from
those whose instructions are sufficiently short. For example, if the cap on instruction length is 5, we would exclude any unit that is produced by a building
with a prerequisite chain exceeding a depth of 5. If the resulting instruction had length 3, we would repeat this process again but for instructions of length 2.
We iterate these steps until the instruction reaches the desired length or no
valid instructions are possible.

\subsection{Spawning of world objects}
Each episode begins with a Nexus building (per the original game) and three Probe workers.
We choose the number of other initial buildings at random between 0 and 36 (the number of  grids in our $6\times6$ environment. We choose the starting location of all buildings
uniformly at random (although no two buildings are permitted to occupy the same grid).
The three Probe workers spawn at the Nexus (per the original game).

\section{Minecraft environment details}
\subsection{Instruction grammar}

\begin{grammar}

    <task> ::= \{ `,' <line-type> \}

    <line-type> ::= <subtask> | <while-expression> | <if-expression>

    <while-expression> ::= while <object> do <subtask-list>

    <if-expression> ::= <if-block> <else-block> | <if-block>

    <if-block> ::= if <object> do <subtask-list>

    <else-block> ::= else do <subtask-list>

    <subtask-list> ::= \{ `,' <subtask> \}

    <subtask> ::= <interaction> | <resource>

    <interaction> ::= inspect | pickup | transform

    <resource> ::= iron | gold | wood

\end{grammar}
\subsection{Generation of instructions}
\label{minecraft:instructions}
Tasks are generated randomly. Most lines in the task are sampled uniformly at
random from \{\emph{If}, \emph{While}, \emph{Subtask}\}. If the current line is inside an if-clause and
the preceding line is a subtask, then \{\emph{Else}, \emph{EndIf}\} is added to the list of
randomly sampled line types. Similar rules apply to while-clauses (we add
\emph{EndWhile} to the list) and else-clauses (we add \emph{EndIf}).

\subsection{Spawning of world objects}
At the beginning of each episode we choose the number $n$ of resources/merchants
uniformly at random between 0 and 36 (the number of grids in our
$6\times6$ griworld). We then sample uniformly at random $n$ times  from $\set{\emph{iron}, \emph{gold},
        \emph{wood}, \emph{merchant}}$ to select the candidate population of the
gridworld. We then test the feasibility of the environment
for the instruction by checking that the requisite resources exist for each
subtask that the agent will have to perform. If the environment is deemed
infeasible, we perform the aforementioned sampling process again. After 50
resamples, if we have still failed to generate a feasible environment, we
generate a new instruction per section
\ref{minecraft:instructions}.
Once we have
generated a feasible population, if $n \le 30$,
we place
water in a straight line through a random index and at a random horizontal/vertical orientation.
Next we place the remaining $n$ resources/merchants and the agent each in unique,
random, open grids.
At this point, we check that the agent has access to a wood resource (with which to build a
bridge), and if not, we remove the water from the map.
Finally, we place walls at any open even-indexed tiles (this to
ensure that walls do not cut off parts of the gridworld).

\begin{figure*}[t]
    \begin{subfigure}[t]{.66\columnwidth}
        \includegraphics[width=\columnwidth]{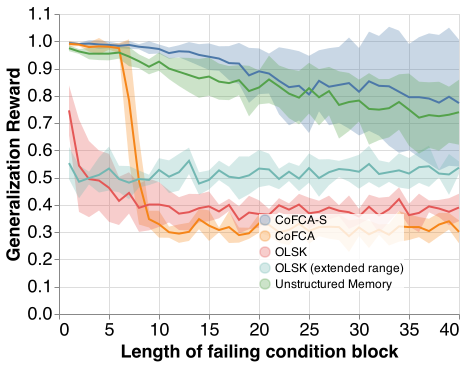}
    \caption{Generalization by condition block size. X-axis corresponds to instruction length; Y-axis is mean success per episode.}
    \label{fig:long-jump}
    \end{subfigure}%
    \hfill
    \begin{subfigure}[t]{.66\columnwidth}
        \includegraphics[width=\columnwidth]{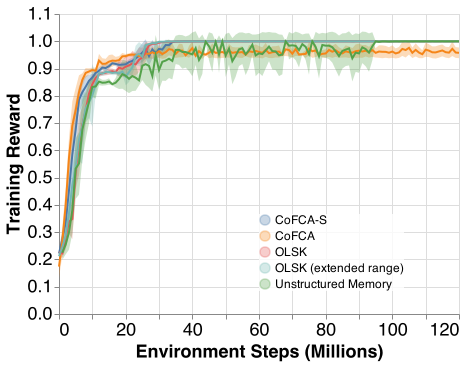}
    \caption{Cumulative reward on training episodes for StarCraft environment}
        \label{fig:starcraft-train}
    \end{subfigure}%
    \hfill
    \begin{subfigure}[t]{.66\columnwidth}
        \includegraphics[width=\columnwidth]{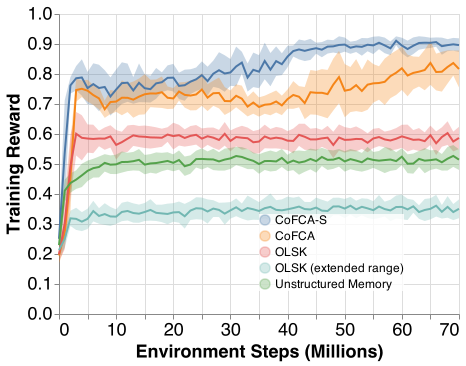}
    \caption{Cumulative reward on training episodes for Minecraft environment}
        \label{fig:minecraft-train}
    \end{subfigure}%
    \label{subfig:train}
    \label{fig:length}
\end{figure*}%

\section{Analysis of long pointer movements in the Minecraft domain}

Here we present results assessing the agents' capability to perform larger pointer
movements than those learned during training. 
We trained the agents on instructions from the Minecraft domain with lengths
sampled randomly from between 1 and 10 (the same training regimen as in   \S\ref{exp:instruction-length}).
We evaluated the agent on a special instruction beginning with a failing condition-block (\emph{if} or \emph{while})
that extends to the end of the instruction, followed by a concluding subtask. 
Thus a successful agent will generally have to jump over the failing condition block
to reach the final subtask.
We varied the length of the failing condition block between 1 and 40 and noted each agent's
performance at each length. These results are shown in
Figure \ref{fig:long-jump}.  This experiment identifies one of the key failure points of
the \name{} architecture that the Scan mechanism is intended to address.
The \name{} architecture is unlikely to sample pointer
movements larger than those it was trained to perform. 
 Concretely, if the agent has never seen a control-flow
block larger than $n$, and $\edges$ has always placed zero mass on pointer movements
greater than $\pm n$, it is unlikely that it will ever place more than zero mass
on those movements, even when longer control-flow blocks require them.
This explains the precipitous
 drop in its performance in Figure \ref{fig:long-jump} as soon as the required jump
 exceeds the largest that it might have encountered in its training set. We also
 note the relatively strong performance of the unstructured memory
 architecture, comparable to CoFCA-S; this shows that the recurrence within the unstructured memory is able to handle longer condition blocks but is unable to deal with multiple control flows accounting for its relatively poor performance in the other generalization experiments. 
 Finally, we note that OLSK (extended range) maintains consistently poor performance irrespective of the condition-block length 
 because, as noted in \S\ref{minecraft-main-results}, this architecture ignores the instruction, having never learned to interpret it in the first place.

\section{Discussion of training performance}
Figures \ref{fig:starcraft-train} and \ref{fig:minecraft-train} display training performance on the StarCraft and Minecraft domain. Training performance for the 
baselines was lower on the Minecraft domain because it
requires more fine-grained control of the pointer. Results do not in any way
compensate for the failure buffer discussed in \S\ref{failure-buffer} and that 
mechanism therefore depresses the performance of the algorithms. On the Minecraft domain, none
of the baselines learned to consistently sequence substasks for longer instructions.

\section{Pseudocode / schematics for baselines}
This section provides pseudocode and schematics for our baselines. We indicate sections
that deviate from the algorithms given in Fig.~\ref{schematic} with red
highlighting. In these sections, we retain the variable names given in Section
~\ref{approach}. For review:
\begin{itemize}
    \item $\mem$: an encoding of the instructions.
    \item $\ptr{t}$: the integer pointer into $\mem$.
    \item $\policy$: the policy, implemented as a neural network.
    \item $\obs{t}$: the observation for the current time-step.
    \item $\edges$: a collection of possible pointer movement distributions.
    \item $\edgeNet{}$: a neural network that chooses among these distributions.
    \item $\gate{t}$: a binary value that permits or prevents movement of $\ptr{t}$.
    \item $\gateNet{}$: a neural network that determines the value of the gate value.
\end{itemize}

\label{baselines}

\begin{algorithm}
    \caption*{Unstructured Memory}
    \begin{algorithmic}[1]
        \STATE $\mem \gets \bagOfWords{\task}$
        \highlight{
            \STATE initialize $\hidden{0} \in \realspace{\hiddenDim}$
            \STATE $\bigruOutput \gets \bigruOf{\mem}$
            \COMMENT{running from first to last index of $\instruction$}}
        \FOR{time step $t$ in episode}
        \STATE $\action{t} \sim \policyOf{\obs{t}, \memi{\ptr{t}}}$
        \STATE $\highlight{\hidden{t}}  \gets \edgeNetOf{\obs{t},
                \highlight{\bigruOutput,} \action{t}, \highlight{\hidden{t-1}}}$
        \STATE $\gate{t} \sim \gateNetOf{\obs{t}, \hidden{t}, \action{t}}$
        \ENDFOR
    \end{algorithmic}
\end{algorithm}

\begin{algorithm}
    \caption*{OLSK}
    \begin{algorithmic}[1]
        \STATE $\ptr{0} \gets 0$
        \STATE $\mem \gets \bagOfWords{\task}$
        \highlight{
            \STATE initialize $\hidden{0} \in \realspace{\hiddenDim}$}
        \FOR{time step $t$ in episode}
        \STATE $\action{t} \sim \policyOf{\obs{t}, \memi{\ptr{t}}}$
        \STATE $\edgeChoice{t}, \highlight{\hidden{t}}  \gets \edgeNetOf{\obs{t},
                \memi{\ptr{t}}, \action{t}, \highlight{\hidden{t-1}}}$
        \highlight{\COMMENT{$\edgeChoice{t} \in \realspace{3}$}}
        \STATE $\edgeChoiceSoftmax{t} \gets \softmax{\edgeChoice{t}}$
        \STATE $\ptrDelta{t} \sim \cat{\edgeChoiceSoftmax{t}}$
        \STATE $\gate{t} \sim \gateNetOf{\obs{t}, \memi{\ptr{t}}, \action{t}}$
        \STATE $\ptr{t + 1} \gets \ptr{t} + \gate{t}\ptrDelta{t}$
        \ENDFOR
    \end{algorithmic}
\end{algorithm}

\begin{algorithm}
    \caption*{OLSK with extended range}
    \begin{algorithmic}[1]
        \STATE $\ptr{0} \gets 0$
        \STATE $\mem \gets \bagOfWords{\task}$
        \highlight{
            \STATE initialize $\hidden{0} \in \realspace{\hiddenDim}$}
        \FOR{time step $t$ in episode}
        \STATE $\action{t} \sim \policyOf{\obs{t}, \memi{\ptr{t}}}$
        \STATE $\edgeChoice{t}, \highlight{\hidden{t}}  \gets \edgeNetOf{\obs{t},
                \memi{\ptr{t}}, \action{t}, \highlight{\hidden{t-1}}}$
        \highlight{\COMMENT{$\edgeChoice{t} \in \realspace{2\instructionLength}$}}
        \STATE $\edgeChoiceSoftmax{t} \gets \softmax{\edgeChoice{t}}$
        \STATE $\ptrDelta{t} \sim \cat{\edgeChoiceSoftmax{t}}$
        \STATE $\gate{t} \sim \gateNetOf{\obs{t}, \memi{\ptr{t}}, \action{t}}$
        \STATE $\ptr{t + 1} \gets \ptr{t} + \gate{t}\ptrDelta{t}$
        \ENDFOR
    \end{algorithmic}
\end{algorithm}

\begin{algorithm}
    \caption*{\name{}}
    \begin{algorithmic}[1]
        \STATE $\ptr{0} \gets 0$
        \STATE $\mem \gets \bagOfWords{\task}$
        \FOR{time step $t$ in episode}
        \STATE $\bigruOutput \gets \bigruOf{\mem, \obs{t}}$
        \highlight{
            \COMMENT{Here $\bigruOutput$ refers to the \emph{last} output of $\bigruOf{\mem}$}
        }
        \STATE \highlight{$\edges \gets \xi \left(\bigruOutput\right)$ }
        \highlight{
            \COMMENT{ $\xi$ is a linear projection}
        }
        \STATE $\action{t} \sim \policyOf{\obs{t}, \memi{\ptr{t}}}$
        \STATE $\edgeChoice{t}  \gets \edgeNetOf{\obs{t},
                \memi{\ptr{t}}, \action{t}}$
        \STATE $\edgeChoiceSoftmax{t} \gets \softmax{\edgeChoice{t}}$
        \STATE $\ptrDelta{t} \sim \cat{\edges\edgeChoiceSoftmax{t}}$
        \STATE $\gate{t} \sim \gateNetOf{\obs{t}, \memi{\ptr{t}}, \action{t}}$
        \STATE $\ptr{t + 1} \gets \ptr{t} + \gate{t}\ptrDelta{t}$
        \ENDFOR
    \end{algorithmic}
\end{algorithm}


\newcommand{\fwidth}{.7}
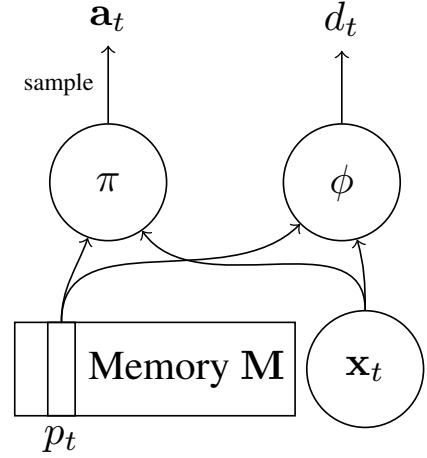
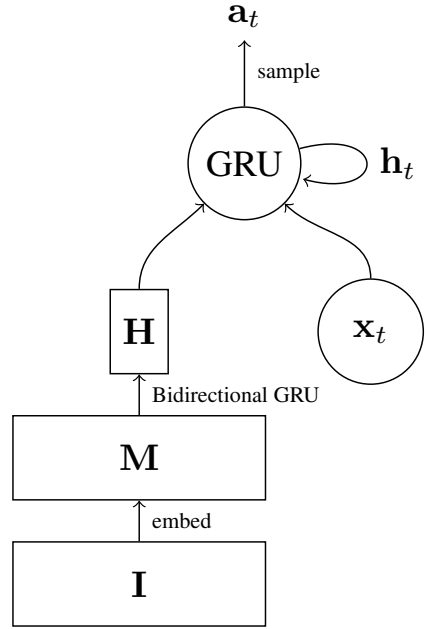
\begin{figure}
     \centering

     \begin{subfigure}[b]{\fwidth\columnwidth}

        \resizebox*{\textwidth}{!}{

\begin{tikzpicture}[x=2mm,y=2mm,
        bin/.style={
                rectangle,
                minimum height=8mm,
                thin, draw,
            },
        weight/.style={
                rectangle,
                minimum height=1mm,
                thin, draw,
            },
        weights/.style={
                rectangle,
                minimum height=1mm,
                minimum width=24mm,
                thin, draw,
            },
        memory/.style={
                rectangle,
                minimum height=8mm,
                minimum width=24mm,
                thin, draw,
            },
    ]
    \node[memory,label={[xshift=-19.0mm]right:Memory $\mem$}]
    (M) at (-3, 2) {};
    \node at (-7, -1) {$\ptr{t}$};
    \node[bin]                            (Mp) at (-7, 2) {};
    \node[minimum size=1cm, draw,circle]  (x) at (6, 2) {$\obs{t}$};
    \node[draw,circle,minimum size=1cm]   (edgeNet) at (5, 10) {$\edgeNet$};
    \node[draw,circle,minimum size=1cm]   (upperPolicy) at (-5, 10) {$\policy$};
    \node                                 (p1) at (5, 17) {$\ptrDelta{t}$};
    \node                                 (a) at (-5, 17) {$\action{t}$};


    \draw [->] (Mp)          to [out=90,in=225] (edgeNet);
    \draw [->] (Mp)          to [out=90,in=250] (upperPolicy);
    \draw [->] (x)          to  [out=90, in=285] (edgeNet);
    \draw [->] (x)          to  [out=90,in=305] (upperPolicy);
    \draw [->] (upperPolicy) to  node[left] {\tiny sample} (a);
    \draw [->] (edgeNet) to (p1);

\end{tikzpicture}
}
\caption{Schematic of OLSK / OLSK (extended-range).}
\end{subfigure}
\\
     \begin{subfigure}[b]{\fwidth\columnwidth}
        \resizebox*{\textwidth}{!}{
\begin{tikzpicture}[x=2mm,y=2mm,
        bin/.style={
                rectangle,
                minimum height=8mm,
                thin, draw,
            },
        memory/.style={
                rectangle,
                minimum height=8mm,
                minimum width=24mm,
                thin, draw,
            },
    ]
    \node[memory] (I) at (-5, -10) {$\instruction$};
    \node[memory] (M) at (-5, -4) {$\mem$};
    \node[bin] (H) at (-5, 2) {$\bigruOutput$};
    \node[minimum size=1cm, draw,circle]  (x) at (6, 2) {$\obs{t}$};
    \node                                 (a) at (0, 17) {$\action{t}$};
    \node[draw,circle,minimum size=1cm]   (GRU) at (0, 10) {GRU};
    \draw [->] (I)  -> node[right] {\tiny embed} (M);
    \draw [->] (M)  -> node[right] {\tiny Bidirectional GRU} (H);
    \draw [->] (H)  to [out=90,in=225] (GRU);
    \draw [->] (x)   to [out=90,in=315] (GRU);
    \draw [->] (GRU) -> node[right] {\tiny sample} (a);
    \draw [->] (GRU) edge[loop right] node {$\hidden{t}$} (h);

\end{tikzpicture}
}
\caption{Schematic of Unstructured Memory.}
\end{subfigure}
\caption{ Schematics for baselines. Note that the schematic for \name{} does not differ from \name{}-S and is therefore omitted.}
\end{figure}
\label{no-pointer}

\setlength{\grammarparsep}{20pt plus 1pt minus 1pt} 
\twocolumn[
    \section{Hyperparameters}
    \subsection{StarCraft}
    \begin{tabular}{l|lllll}
                                       & \name{}-S & \name{} & \vtop{\hbox{\strut Unstructured}\hbox{\strut Memory}} & OLSK & OLSK-E
        \\ \hline
        convolution hidden sizes       & 250       & 250     & 150                                                   & 250 & 250
        \\
        convolution kernel sizes       & 2         & 2       & 2                                                     & 2 & 2
        \\
        convolution strides            & 1         & 1       & 1                                                     & 1 & 1
        \\
        $\edgeNet$ hidden size         & 250       & 250     & 200                                                   & 200 & 200
        \\
        $\embeddingDim$                & 200       & 100     & 100                                                   & 150 & 150
        \\
        entropy coefficient            & 0.02      & 0.02    & 0.02                                                  & 0.02 & 0.02
        \\
        learning rate                  & 8e-5      & 7.5e-5  & 4e-                                                   & 4e-05 & 4e-05
        \\
        $\numEdges$                    & 3         & 2       & NA                                                    & NA & NA
        \\
        time steps per gradient update & 35        & 30      & 35                                                    & 30 & 30
        \\
        gradient steps per update      & 6         & 7       & 7                                                     & 7 & 7
        \\
    \end{tabular}
    \vspace{1cm}
    \subsection{Minecraft}
    \begin{tabular}{l|lllll}
                                       & \name{}-S & \name{} & \vtop{\hbox{\strut Unstructured}\hbox{\strut Memory}} & OLSK & OLSK-E
        \\ \hline
        convolution hidden sizes       & 32,32     & 64,32   & 32,32                                                 & 64,16 & 64,16
        \\
        convolution kernel sizes       & 2,2       & 2,2     & 2,2                                                   & 2,2 & 2,2
        \\
        convolution strides            & 2,2       & 2,2     & 2,2                                                   & 2,2 & 2,2
        \\
        $\edgeNet$ hidden size         & 128       & 128     & 64                                                    & 64 & 64
        \\
        $\embeddingDim$                & 64        & 32      & 64                                                    & 32 & 32
        \\
        entropy coefficient            & 0.015     & 0.015   & 0.015                                                 & 0.015 & 0.015
        \\
        learning rate                  & 0.0025    & 0.0025  & 0.0025                                                & 0.0025 & 0.0025
        \\
        $\numEdges$                    & 2         & 9       & NA                                                    & NA & NA
        \\
        time steps per gradient update & 25        & 25      & 25                                                    & 25 & 25
        \\
        gradient steps per update      & 2         & 2       & 2                                                     & 2 & 2
        \\
    \end{tabular}
]

\end{document}